\documentclass[letterpaper]{article} 
\usepackage[preprint]{aaai2027}  
\usepackage[hyphens]{url}  
\usepackage{graphicx} 
\usepackage{natbib}  
\usepackage{caption} 
\usepackage{algorithm}
\usepackage{algorithmic}

\usepackage{newfloat}
\usepackage{listings}
\DeclareCaptionStyle{ruled}{labelfont=normalfont,labelsep=colon,strut=off} 
\lstdefinestyle{artifactcontent}{%
	basicstyle={\fontsize{9}{10}\selectfont\ttfamily},
	numbers=none,
	xleftmargin=0pt,
	frame=single,
	framerule=0.45pt,
	rulecolor=\color{black!50},
	backgroundcolor=\color{white},
	framesep=7pt,
	aboveskip=0pt,
	belowskip=0pt,
	showstringspaces=false,
	keepspaces=true,
	columns=fullflexible,
	tabsize=2,
	breaklines=true}
\floatstyle{ruled}
\newfloat{listing}{tb}{lst}{}
\floatname{listing}{Listing}

\usepackage{booktabs}
\usepackage{multirow}
\usepackage{tabularx}
\newcolumntype{Y}[1]{>{\hsize=#1\hsize\linewidth=\hsize\centering\arraybackslash}X}
\usepackage{amssymb}
\usepackage{xcolor}

\title{MerchantBench: Benchmarking LLM Agents for Long-Term Coherence in E-Commerce Operations}
\author{
\normalsize
Qiming Shi\textsuperscript{2}, Yulong Tao\textsuperscript{1}, Linbo Jin\textsuperscript{1}\footnotemark[1], Zhaolu Kang\textsuperscript{4}, Yibo Dou\textsuperscript{4},\\
Jiawen Zhu\textsuperscript{3}, Tianjun Pan\textsuperscript{5}, Shaokang Fu\textsuperscript{1}, Chengyu Wang\textsuperscript{1},\\
Siyue Li\textsuperscript{1}, Yaping Cheng\textsuperscript{1}, Di Weng\textsuperscript{3}\thanks{Corresponding authors.}, Chengfu Huo\textsuperscript{1}
}
\affiliations{
\textsuperscript{1}Alibaba Group\\
\textsuperscript{2}State Key Lab of CAD\&CG, Zhejiang University\\
\textsuperscript{3}School of Software Technology, Zhejiang University\\
\textsuperscript{4}School of Software and Microelectronics, Peking University\\
\textsuperscript{5}College of Computer Science and Artificial Intelligence, Fudan University
}

\begin{document}

\maketitle

\begin{abstract}
Large language model agents are increasingly evaluated as autonomous tool users, yet most benchmarks focus on bounded tasks with immediate success criteria.
Real-world deployments often require Long-Term Coherence, the capacity to preserve purposeful behavior across extended horizons while adapting decisions to accumulated evidence.
Evaluating this capacity requires a persistent environment in which actions constrain future choices, feedback arrives at heterogeneous delays, and incoherent behavior produces measurable cumulative effects.
Seller-side e-commerce provides a suitable setting for this evaluation through recurrent and interdependent decisions over Product Sourcing, Listing and Pricing Control, Cash-Flow Management, and Mixed-Latency Feedback Adaptation.
We introduce MerchantBench, a 365-day order-level simulation grounded in 98,843 real e-commerce product records and equipped with 26 tools for agent interaction.
MerchantBench couples promptly observable Upstream Supplier Events with delayed Downstream Order Outcomes, requiring agents to follow individual order lifecycles and revisit earlier decisions.
We evaluate eight LLMs under two agent frameworks in 48 runs, each spanning 365 simulated days.
Our results reveal a substantial gap between even the latest LLMs and human participants, with the best LLM configuration attaining only 27.3\% of the mean final net assets achieved by human participants.
Our code is available at \url{https://github.com/KhanCold/merchantbench}.
\end{abstract}

\section{Introduction}

Large language models have become a foundation for autonomous agents that plan, call tools, interact with external systems, and make decisions over multiple turns.
General agent benchmarks now measure tool use, web interaction, application control, and state-changing workflows \cite{agentbench2024,webarena2024,taubench2024,appworld2024}.
Yet many real-world deployments are not bounded tasks with immediate and unambiguous completion criteria.
They require agents to operate over extended horizons in environments whose state persists, where earlier actions constrain later options and relevant consequences may emerge only after many intervening decisions.
Evidence from recent benchmarks indicates that even state-of-the-art agents often fail to maintain coherent performance over long horizons \cite{ultrahorizon2025,odysseybench2025,osworld22026}.
These results motivate extending agent evaluation beyond isolated task completion to examine sustained objective pursuit and decision consistency in realistic long-horizon environments.

\begin{figure}[t]
\centering
\includegraphics[width=\columnwidth]{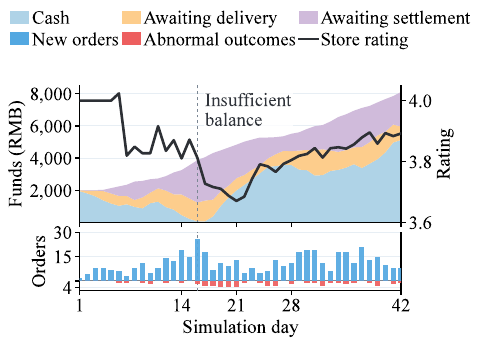}
\caption{Order-level dynamics couple immediate liquidity pressure with delayed feedback.
New orders commit available cash before settlement, whereas abnormal outcomes surface later and affect store rating.}
\label{fig:order_risk_propagation}
\end{figure}

Seller-side e-commerce provides a suitable setting for evaluating Long-Term Coherence.
Unlike bounded tasks with explicit completion criteria, online store operation requires continued intervention throughout the operating horizon.
The agent must repeatedly select products, control listings and prices, manage limited cash, and revise earlier decisions as market conditions, supplier states, and order outcomes evolve.
Seller-side e-commerce therefore tests whether an agent can sustain and revise a merchant policy over time, rather than complete an isolated action.

Vending-Bench and RetailBench have taken important steps toward realistic long-horizon business evaluation \cite{vendingbench2025,retailbench2026}.
Vending-Bench evaluates vending-machine operation, while RetailBench models supermarket management over a fixed catalog of 96 grocery products.
Seller-side e-commerce, however, introduces two challenges that are not represented together in these environments.

First, e-commerce feedback is generated through individual order lifecycles.
A listing or pricing decision may create new orders and commit available cash immediately, whereas fulfillment failures and after-sales outcomes become observable only later.
Figure~\ref{fig:order_risk_propagation} visualizes this temporal asymmetry between immediate order-driven cash commitments and delayed abnormal outcomes.
As delayed evidence accumulates, the agent must associate later outcomes with earlier decisions and determine whether its current merchant policy should be maintained or revised.

Second, a long episode alone does not meaningfully test Long-Term Coherence if the available products and their demand remain static.
A large, data-grounded Product Catalog with full-year demand trajectories creates a changing opportunity set in which promising products emerge and existing choices lose value over time.
The agent must therefore continually identify new opportunities and revise its portfolio using market signals and realized order outcomes.

To address this gap, we introduce MerchantBench, which evaluates Long-Term Coherence through persistent seller-side e-commerce operation.
MerchantBench formulates e-commerce operation as a partially observable decision-making problem over 365 simulated days.
The environment grounds a Product Catalog in 98,843 real e-commerce product records and converts demand into individual orders that progress through fulfillment and after-sales stages.
Through merchant-visible tools, the agent performs Product Sourcing, controls listings and prices, manages cash flow, and monitors supplier and order states.
Upstream Supplier Events and Downstream Order Outcomes become observable at different times, requiring the agent to use later evidence to maintain or revise earlier decisions.
Across the operating horizon, the simulator updates demand, supplier states, order lifecycles, cash flow, penalties, and store reputation.

Our contributions are as follows:
\begin{itemize}
    \item To our knowledge, we introduce the first benchmark for evaluating Long-Term Coherence through persistent seller-side e-commerce operation, structured around four interdependent decision components.
    \item We develop an order-level simulation environment grounded in 98,843 real e-commerce product records, with partial observability, cash constraints, Upstream Supplier Events, and delayed Downstream Order Outcomes.
    \item We conduct 48 runs of 365 simulated days across eight LLMs and two agent frameworks and analyze business outcomes and decision traces to characterize their Long-Term Coherence.
\end{itemize}

\section{Related Work}

\paragraph{Agent Evaluation in Commerce.}
Existing benchmarks cover shopping and storefront interaction \cite{webshop2022,shoppingbench2026,amazonbench2026,shopsimulator2026,shopgym2026,ecomagentbench2026}, customer support \cite{ecombench2025}, and merchant workflows and negotiation \cite{ecomstage2026,bargainbench2025}.
Market-Bench and Magentic Marketplace examine market competition and transactions among economic agents \cite{marketbench2026,magenticmarketplace2025}.
Across these categories, evaluation centers on tasks, dialogues, transactions, or competitive episodes rather than continuous operation of the same online store.

\paragraph{Long-Horizon Agent Evaluation.}
Recent long-horizon benchmarks assess sustained reasoning and action across workplace workflows, open-ended exploration, virtual-world planning, web navigation, computer use, inventory control, order fulfillment, and interactive economies \cite{ultrahorizon2025,odysseybench2025,herobench2025,odysseys2026,wildclawbench2026,osworld22026,inventorybench2026,ofcourse2023,ecogym2026,coffeebench2026}.
Vending-Bench and Vending-Bench 2 emphasize sustained business operation, while RetailBench evaluates evidence acquisition, action conversion, and temporal follow-up in supermarket management \cite{vendingbench2025,vendingbench22025,retailbench2026}.
MerchantBench extends this line by evaluating how agents adapt to delayed order-level feedback and nonstationary demand derived from real e-commerce data.

\section{MerchantBench}

\begin{figure*}[t]
\centering
\includegraphics[width=\textwidth]{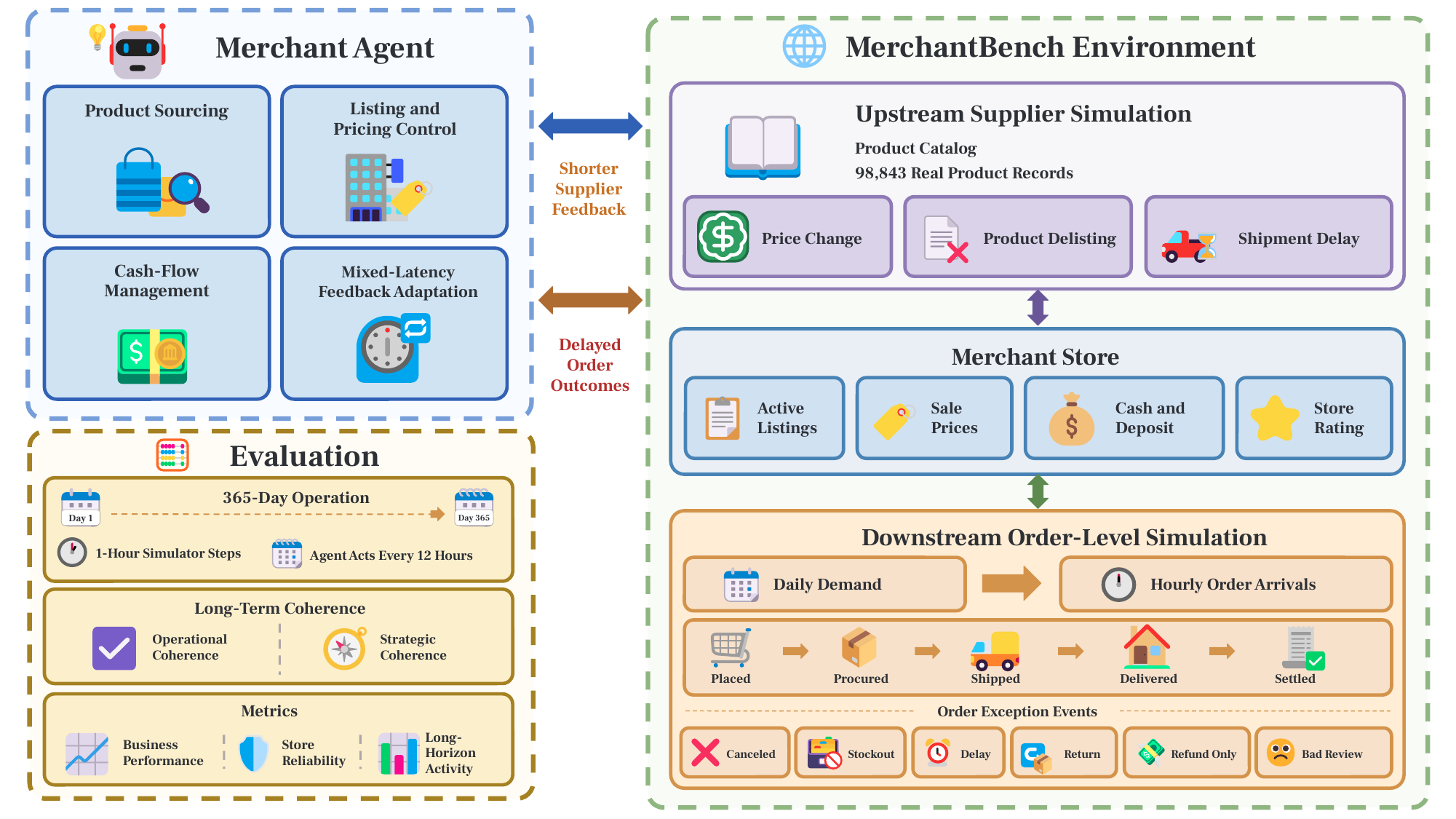}
\caption{Overview of MerchantBench.
The merchant agent coordinates four decision components through shorter supplier feedback and delayed order outcomes over 365 days.
The environment combines an upstream supplier simulation, a merchant store, and a downstream order level simulation to evaluate Long-Term Coherence and business outcomes.}
\label{fig:merchantbench_method}
\end{figure*}

\subsection{Task Formulation}

We formulate store operation as a finite horizon partially observable Markov decision process (POMDP) \cite{kaelbling1998pomdp}
\begin{equation}
\mathcal{M}=\langle\mathcal{S},\mathcal{A},P,\mathcal{O},Z,R,\mu_0,H_c\rangle.
\label{eq:pomdp}
\end{equation}
The simulator advances hourly over a 365 day control horizon, giving $H_c=8{,}760$ steps indexed by $t\in\{0,\ldots,H_c-1\}$.
Demand, supplier states, and order lifecycles evolve at every step, while the agent receives a decision window once every 12 steps.
The latent state $s_t\in\mathcal{S}$ contains the simulation clock, product demand profiles, supplier conditions, store listings and finances, active orders, and pending events.
The initial state follows $\mu_0$ and includes the cash balance, security deposit, and listing capacity.
The cash balance funds procurement and realized losses, unpaid fines draw from the security deposit, and operation terminates when the deposit is exhausted.
At activation steps, $a_t\in\mathcal{A}$ denotes the sequence of merchant tool invocations within the decision window, while other steps use a fixed null action.
The transition kernel $P(s_{t+1}\mid s_t,a_t)$ combines tool induced store changes with autonomous demand, supplier, and order evolution, with listing and pricing changes affecting demand from the next step.
The observation kernel $Z(o_{t+1}\mid s_{t+1},a_t)$ exposes only merchant visible information, while demand profiles, risk parameters, pending outcomes, and future event times remain latent until their effects become observable.
The policy therefore conditions on observation and tool result history rather than the full state.
At $t=H_c$, new demand and agent activations stop while active orders continue until terminal settlement at a terminal step $T\geq H_c$.
Intermediate rewards are zero, and the objective is expected terminal net assets
\begin{equation}
\begin{array}{rcl}
J(\pi)&=&\mathbb{E}_{\pi}[R(s_T)],\\
R(s_T)&=&B_T+D_T+I_T+Q_T.
\end{array}
\label{eq:terminal_reward}
\end{equation}
where $B_T$, $D_T$, $I_T$, and $Q_T$ denote the terminal cash balance, security deposit, funds in transit, and receivables.
Thus, $R(s_T)$ is the realized net asset value of one run and $J(\pi)$ is its expectation across stochastic trajectories.

\subsection{Real-World Data Grounding}

\begin{figure}[h]
\centering
\includegraphics[width=\columnwidth]{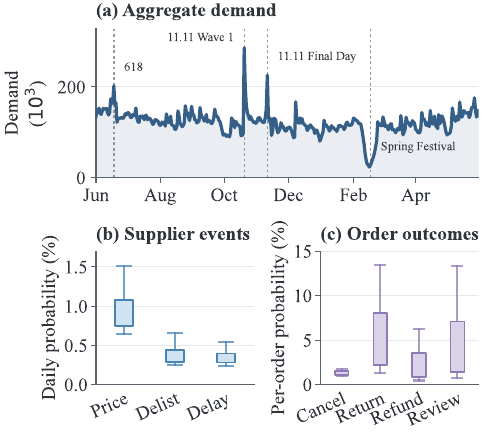}
\caption{Real-world demand patterns and calibrated risk profiles across 98,843 products.
Whiskers span the 10th to 90th percentiles, and boxes show interquartile ranges.}
\label{fig:data_environment_grounding}
\end{figure}

MerchantBench is grounded in real-world e-commerce data from 1688, the largest integrated domestic wholesale marketplace in China \cite{alibaba1688}.
The data contain product and supplier attributes, 365-day product-level demand histories, and platform quality and fulfillment signals.
The data cover 365 days from June 1, 2025 through May 31, 2026.
Alongside the product data, MerchantBench incorporates 365 daily market reports from 1688 as date-aligned signals for product sourcing.
We select 10 first-level product categories spanning apparel, household and office goods, appliances, pet and gardening products, toys, bags, and sports and outdoor products.
After excluding records with missing identifiers, unmapped categories, nonpositive prices, or incomplete demand histories, the dataset contains 98,843 products from 36,576 suppliers.
Through catalog and supplier tools, the agent accesses only public product and supplier attributes.
Figure~\ref{fig:data_environment_grounding} captures aggregate demand peaks at 618 and during the first wave and final day of 11.11, together with a trough during the Spring Festival.
The lower panels show the distributions of effective Upstream Supplier Event and Downstream Order Outcome probabilities.

\subsection{Upstream Supplier Simulation}

The supply pool maintains time varying procurement prices, available inventory, availability, and supplier shipment times, while inventory replenishes over time.
At each simulator step, the simulator samples the three Upstream Supplier Events, namely Price Change, Product Delisting, and Shipment Delay, using product level probabilities calibrated from real platform fulfillment signals.
These events alter the upstream procurement price, suspend product procurement, and extend supplier dispatch time, respectively.
Inventory stockouts arise endogenously when incoming orders deplete stock faster than it replenishes.
To prevent persistent environment drift, each triggered abnormality receives a sampled recovery time at which the affected supplier attributes return to their base states.
The agent can observe realized changes to price, availability, quantity, and shipment time through catalog and supplier queries, but it cannot access the underlying abnormality flag, trigger probability, or recovery schedule.

\subsection{Downstream Order Simulation}

\paragraph{Order Level Simulation.}
The downstream simulation converts product level daily demand traces into individual orders.
For product $i$ listed by merchant $m$ at time $t$, the hourly arrival intensity is
\begin{equation}
\lambda_{m,i,t}=D_{i,d(t)}w_{c(i),h(t)}r_{m,t}\ell_{m,i,t}
\left(p_{m,i,t}/p_i^{\mathrm{ref}}\right)^{-\epsilon_i}.
\label{eq:order_intensity}
\end{equation}
The indices $d(t)=\lfloor t/24\rfloor$ and $h(t)=t\bmod 24$ denote the data day and hour of day associated with step $t$.
The quantity $D_{i,d(t)}$ is the linked daily demand from the real-world data, and $w_{c(i),h(t)}$ distributes category demand across hours.
The price term uses product elasticity $\epsilon_i$, while $r_{m,t}$ and $\ell_{m,i,t}$ capture store rating and listing exposure.
Realized order outcomes update the published store rating after each completed day, and its discrete star level determines $r_{m,t}$.
The factor $\ell_{m,i,t}$ increases during a listing's cold start and then decays with age.
The appendix provides precise definitions of both factors in the section on demand and rating dynamics.
The environment samples $N_{m,i,t}\sim\mathrm{Poisson}(\lambda_{m,i,t})$ and instantiates each arrival as an order candidate.
At creation, each candidate receives one latent customer outcome from normal fulfillment, Cancellation, Returnless Refund, Return and Refund, or Bad Review according to its product specific risk profile.
Stockout and Late Shipment instead arise from procurement and fulfillment dynamics.
The selected outcome and its realization time remain hidden until the corresponding lifecycle transition occurs.

\begin{table*}[t]
\centering
\small
\setlength{\tabcolsep}{1.5pt}
\renewcommand{\arraystretch}{1.10}
\renewcommand{\tabularxcolumn}[1]{m{#1}}
\begin{tabularx}{\textwidth}{@{}l Y{1.15} Y{0.80} Y{1.20} Y{0.85} Y{0.70} Y{1.15} Y{1.15} Y{1.15} Y{1.05} Y{0.80}@{}}
\toprule
\multirow{2}{*}[-1.7ex]{\shortstack[l]{\textbf{Model or}\\\textbf{Operator}}}
& \multicolumn{4}{c}{\textbf{Business Performance}}
& \multicolumn{3}{c}{\textbf{Store Reliability}}
& \multicolumn{3}{c}{\textbf{Long-Horizon Activity}} \\
\cmidrule(lr){2-5}\cmidrule(lr){6-8}\cmidrule(lr){9-11}
& \shortstack{\textbf{Net Assets}}
& \shortstack{\textbf{GMV}}
& \shortstack{\textbf{Profit}\\\textbf{Margin}}
& \textbf{Orders}
& \textbf{Fines}
& \shortstack{\textbf{Avg. Store}\\\textbf{Rating}}
& \shortstack{\textbf{Anomaly}\\\textbf{Rate}}
& \shortstack{\textbf{Avg. Active}\\\textbf{Listings}}
& \textbf{SWR}
& \shortstack{\textbf{Tool}\\\textbf{Calls}} \\
\midrule
\multicolumn{11}{c}{\textbf{ReAct}} \\
\midrule
GPT-5.6 Sol & \textbf{40.89} & \textbf{74.19} & \textbf{51.3} & 996 & 499 & 4.04 & \textbf{10.7} & \textbf{50.0} & \textbf{99.4} & \textbf{7,257} \\
Claude Opus 4.8 & 31.89 & 69.10 & 44.4 & 1,214 & 796 & \textbf{4.05} & 12.1 & 24.0 & 45.0 & 1,139 \\
Qwen3.7-Max & 20.66 & 39.73 & 44.5 & 925 & 672 & 3.90 & 16.1 & 39.6 & 11.1 & 815 \\
Qwen3.7-Plus & 20.74 & 40.85 & 45.6 & 1,056 & 705 & 3.99 & 13.1 & 49.9 & 52.2 & 1,221 \\
GLM-5.2 & 25.73 & 60.90 & 37.3 & 2,158 & 1,422 & 3.93 & 14.9 & 26.0 & 53.3 & 2,045 \\
DeepSeek-V4-Pro & 6.56 & 8.40 & 41.9 & 450 & \textbf{245} & 4.01 & 14.4 & 23.5 & 30.6 & 660 \\
DeepSeek-V4-Flash & 14.47 & 28.78 & 39.6 & 985 & 517 & 4.04 & 14.1 & 19.3 & 40.6 & 960 \\
Kimi K2.6 & 24.99 & 63.69 & 32.9 & \textbf{2,230} & 1,474 & 3.89 & 15.3 & 47.3 & 10.6 & 1,228 \\
\midrule
\multicolumn{11}{c}{\textbf{Hermes}} \\
\midrule
GPT-5.6 Sol & 52.93 & \textbf{133.07} & 40.2 & 3,251 & 1,096 & \textbf{4.09} & \textbf{9.2} & \textbf{50.0} & \textbf{66.1} & \textbf{4,831} \\
Claude Opus 4.8 & 35.56 & 83.23 & 39.9 & 1,808 & 1,089 & 4.02 & 11.8 & 22.1 & 31.7 & 1,138 \\
Qwen3.7-Max & \textbf{59.46} & 116.76 & 46.9 & 1,929 & 1,295 & 3.90 & 15.7 & 49.6 & 22.2 & 1,366 \\
Qwen3.7-Plus & 29.42 & 53.69 & \textbf{48.9} & 981 & \textbf{642} & 3.95 & 13.8 & 49.9 & 19.4 & 820 \\
GLM-5.2 & 42.32 & 103.06 & 36.9 & 2,731 & 1,454 & 4.05 & 11.3 & 49.6 & 62.8 & 1,792 \\
DeepSeek-V4-Pro & 16.71 & 31.95 & 43.4 & 1,062 & 665 & 3.98 & 14.5 & 33.0 & 33.3 & 942 \\
DeepSeek-V4-Flash & 24.69 & 64.52 & 37.6 & 1,989 & 1,774 & 3.93 & 16.0 & 48.8 & 62.2 & 1,259 \\
Kimi K2.6 & 23.96 & 75.06 & 26.8 & \textbf{3,398} & 2,671 & 3.73 & 19.1 & 48.3 & 17.8 & 969 \\
\midrule
\multicolumn{11}{c}{\textbf{Others}} \\
\midrule
Human & 217.61 & 608.06 & 35.3 & 9,442 & 5,622 & 3.98 & 12.5 & 49.1 & 100.0 & 8,311 \\
Rule-based & 24.48 & 53.37 & 40.3 & 1,605 & 1,374 & 3.76 & 18.0 & 50.0 & 100.0 & 3,236 \\
\bottomrule
\end{tabularx}
\caption{Business performance, store reliability, and long-horizon activity after 365 simulated days.
Values are means over three runs.
Final net assets and GMV are reported in thousands of RMB, total fines in RMB, and rate metrics in percent.
SWR denotes Sustained Window Rate.
The best result within each framework is shown in bold.}
\label{tab:main_results}
\end{table*}

\paragraph{Order Lifecycle.}
MerchantBench follows the single item drop shipping model supported by 1688, in which merchants hold no inventory in advance.
Each Order Placed triggers immediate procurement at the current supplier price, and successful procurement deducts the cash balance, decreases supplier inventory, and moves the order to Procured.
Following supplier and logistics delays, the order advances through Shipped and Delivered, at which point the sale price becomes a receivable.
A normal order becomes Settled after a sampled delay and credits the receivable to the cash balance.

MerchantBench models Normal Fulfillment together with six abnormal Downstream Order Outcomes, namely Cancellation, Stockout, Late Shipment, Returnless Refund, Return and Refund, and Bad Review.
Cancellation restores the procurement cost, while Stockout prevents procurement.
Late Shipment marks a missed dispatch deadline, after which the order continues through fulfillment and settlement.
Both refund outcomes remove the receivable, but only Return and Refund restores the procurement cost, whereas Bad Review preserves the sales revenue.
Stockout, Late Shipment, Return and Refund, and Bad Review incur platform fines.
All abnormal outcomes except Cancellation also contribute adverse evidence to the store rating through outcome-specific experience scores and evidence weights.

\subsection{Agent Interface}

MerchantBench exposes a shared observation protocol and 26 merchant tools, allowing different agent frameworks to interact with identical environment dynamics and observable state.
At each decision window, the agent receives a summary of simulated time, store status, and recent supplier and order changes, then acts until it ends the window or reaches the time limit.
The appendix lists the complete MerchantBench tool interface.

\paragraph{Product Sourcing.}
Daily market reports, catalog search, product details, and public supplier profiles support product selection, while demand, product risk rates, future order outcomes, and future supplier events remain latent.

\paragraph{Listing and Pricing Control.}
Listing, delisting, repricing, and performance views allow agents to construct the store portfolio and revise its products and prices.

\paragraph{Cash-Flow Management.}
Finance and store views expose the cash balance, security deposit, committed funds, expected settlements, fines, and closure conditions, which agents manage through subsequent sourcing, listing, delisting, and pricing decisions.

\paragraph{Mixed-Latency Feedback Adaptation.}
Supplier and order tools reveal Upstream Supplier Events and Downstream Order Outcomes as they unfold, allowing agents to revise the other three decision components in response to mixed-latency feedback.

\section{Experiments}

\subsection{Experimental Setup}

\paragraph{Agent Configurations and Baselines.}
We evaluate eight LLMs under ReAct \cite{react2023} and Hermes \cite{hermesagent2026} with three runs for each pairing of an LLM and a framework.
The evaluated models are GPT-5.6 Sol \cite{gpt562026}, Claude Opus 4.8 \cite{claudeopus482026}, Qwen3.7-Max and Qwen3.7-Plus \cite{qwen372026}, GLM-5.2 \cite{glm522026}, DeepSeek-V4-Pro and DeepSeek-V4-Flash \cite{deepseekv42026}, and Kimi K2.6 \cite{kimik262026}.
ReAct pairs each model with a minimal controller over the 26 MerchantBench tools to assess core planning, reasoning, and tool use.
Hermes uses its default configuration, combining the 26 MerchantBench tools with built in capabilities for code execution, planning, memory, and skill management.
Both frameworks compress long interaction histories, with each evaluated model also serving as its own summarizer.
Each run starts with RMB 2,000 in cash, a RMB 1,000 security deposit, and capacity for 50 active listings.
We additionally compare with a Rule-based baseline and three Human participants without prior e-commerce operating experience.
The Rule-based baseline performs daily checks, removes inactive or supplier-affected products, and fills open listing slots using the daily market report.
The appendix provides the detailed experimental settings.

\paragraph{Evaluation Metrics.}
We evaluate Business Performance using Final Net Assets, GMV, Net Profit Margin, and Orders; Store Reliability using Total Fines, Average Store Rating, and Order Anomaly Rate; and Long-Horizon Activity using Average Active Listings, Sustained Window Rate (SWR), and Total Tool Calls.
SWR is the minimum share of scheduled decision windows containing at least one environment tool call across all rolling 30 day periods.

\subsection{Main Results}

\paragraph{Overall Performance.}
Table~\ref{tab:main_results} reports the final performance of all evaluated configurations and baselines.
GPT-5.6 Sol records the highest final net assets under ReAct, whereas Qwen3.7-Max ranks first under Hermes.
Qwen3.7-Max with Hermes achieves the highest final net assets among all 16 configurations.
When results are aggregated by model across the two frameworks, GPT-5.6 Sol has the highest average final net assets.

\begin{figure*}[t]
\centering
\includegraphics[width=\textwidth]{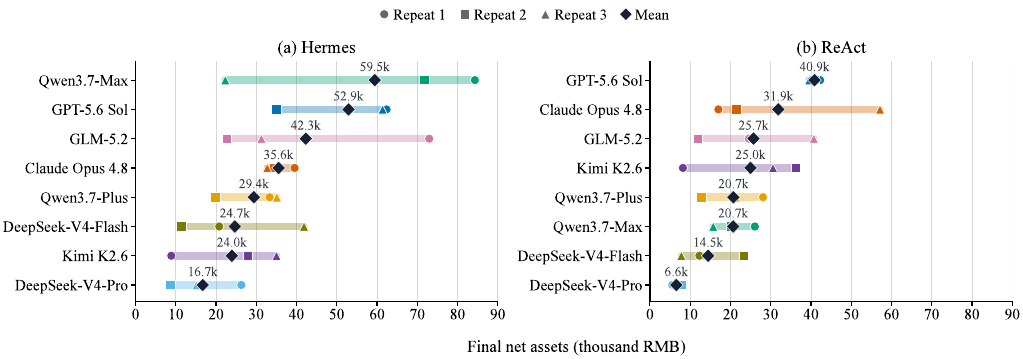}
\caption{Final net asset distributions across three repeated runs for each configuration.}
\label{fig:performance_variability}
\end{figure*}

\paragraph{Performance Variability.}
Figure~\ref{fig:performance_variability} shows substantial differences in stability across configurations.
GPT-5.6 Sol under ReAct and Claude Opus 4.8 under Hermes have the lowest coefficients of variation within their respective frameworks at 3.3\% and 10.0\%.
Despite achieving the highest mean final net assets, Qwen3.7-Max under Hermes is considerably less stable, with a coefficient of variation of 55.1\%.

\paragraph{Framework Analysis.}
Averaged across the eight models, Hermes produces 53.3\% higher final net assets, 71.5\% higher GMV, and 71.2\% more orders than ReAct.
Mean final net assets are higher under Hermes for seven of the eight models, with gains ranging from 11.5\% for Claude Opus 4.8 to 187.8\% for Qwen3.7-Max.
Kimi K2.6 is the sole exception, with mean final net assets 4.1\% lower under Hermes than under ReAct, showing that framework benefits depend strongly on the underlying model.

\subsection{Order-Level Risk Propagation}

Figure~\ref{fig:order_risk_propagation} illustrates the mixed-latency evidence generated by MerchantBench's order-level simulation.
Agents observe prior ratings of upstream catalog products during sourcing and receive prompt demand signals from realized sales, whereas product quality is revealed only through delayed order outcomes that may require product-level risk response or store-level rating adaptation.

\paragraph{Product-Level Risk Response.}
Models differed in whether they converted delayed order outcomes into product-specific interventions.
In representative runs, GPT-5.6 Sol and Kimi K2.6 attributed adverse outcomes to the responsible listing and replaced it, whereas Qwen3.7-Plus retained a risky product for further observation and DeepSeek-V4-Pro did not revise affected listings after refunds.
Human participants described a more complete response for popular but risky products, in which they delisted the product and searched similar keywords for a replacement that could preserve the underlying demand.
These behaviors distinguish simple anomaly detection from the full chain of product attribution, risk removal, and demand-preserving replacement.

\paragraph{Store-Level Rating Adaptation.}
Order-level anomalies also reduce the store rating, allowing product-specific failures to affect demand across the portfolio.
In a GPT-5.6 Sol trajectory, the agent lowered prices on proven products without adverse outcomes to increase normal settlements and recover the rating threshold.
Qwen3.7-Max applied the same mechanism more aggressively by repricing 40 listings after an early bad review reduced the store to three stars.

\subsection{Long-Term Coherence Analysis}

The aggregate results reveal a substantial gap between the evaluated LLM agents and Human operators.
Our trace evidence suggests that two forms of Long-Term Coherence failure developing over extended operation may contribute to this performance gap.
Some agents progressively reduce store intervention and fail to follow up on prior decisions or delayed outcomes, indicating a loss of Operational Coherence.
Others remain active but drift from the terminal net assets objective or fail to revise ineffective policies as evidence accumulates, indicating a loss of Strategic Coherence.

\begin{figure*}[!t]
\centering
\includegraphics[width=\textwidth]{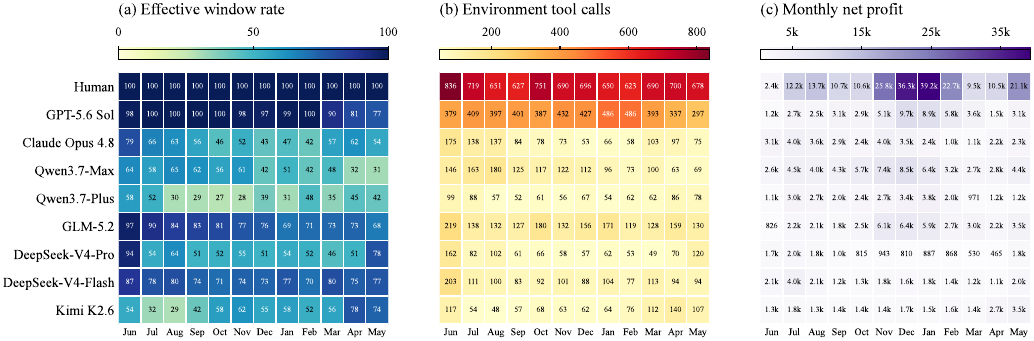}
\caption{Monthly Operational Coherence profiles for the Human baseline and all eight Hermes models.
Panels report effective window rate, environment tool calls, and monthly net profit.}
\label{fig:operational_coherence_selected}
\end{figure*}

\paragraph{Operational Coherence.}
Figure~\ref{fig:operational_coherence_selected} reveals substantial differences in whether merchant activity persists across the operating horizon.
Table~\ref{tab:main_results} shows that Human operators retain an SWR of 100\%, while LLM configurations range from 10.6\% to 99.4\% under ReAct and from 17.8\% to 66.1\% under Hermes.
Qwen3.7-Max provides the clearest contrast, with its quarterly Effective Window Rate falling from 62\% to 37\% under Hermes and more sharply from 68\% to 23\% under ReAct, alongside substantial reductions in environment tool calls.
Similar patterns of Activity Decay appear across several other models under both frameworks.
Monthly net profit shows how business performance evolves alongside these activity patterns.
Such operational decline often originates in strategic drift.

\paragraph{Strategic Coherence.}
Strategic Coherence comprises two complementary dimensions.
Goal Consistency concerns whether decisions across time continue to serve the long-term business objective, whereas Evidence-Calibrated Adaptation concerns how an agent maintains or revises its policy as feedback arrives at different latencies and evidence accumulates over time.

\noindent\textbf{Goal Consistency.}
Goal Consistency fails when agents lose the autonomy to keep pursuing terminal net assets.
Under Control-Loop Narrowing, the sourcing and operating loop gradually collapses into reactive handling of Upstream Supplier Events, with little self-initiated sourcing, repricing, replacement, or diagnosis.
For ReAct Qwen3.7-Max, an SWR of 11.1\% coincides with supply chain checks rising from 14\% to 34\% of its remaining tool calls.
Related activity decay also appears in Claude Opus 4.8, GLM-5.2, DeepSeek-V4-Pro, and Qwen3.7-Plus under Hermes.
At the extreme, Premature Abandonment occurs when an agent concludes that the store cannot recover although feasible actions remain.
In one Hermes Kimi K2.6 run, the agent made this judgment on Day 104 and then took no environment action in 355 of the remaining 523 decision windows.
In both cases, the agent waits for external events or time to change the store instead of operating it autonomously.

\noindent\textbf{Evidence-Calibrated Adaptation.}
Evidence-Calibrated Adaptation examines whether agents revise their policies as liquidity, seasonal demand, and accumulated experience change.

Under initial liquidity constraints, Human and agents operate within similar low price ranges.
As liquidity increases, Human operators broaden the procurement price range and selectively return to lower price and higher throughput products when higher value experiments underperform.
Their mean active listing procurement prices increase from between RMB~43.4 and RMB~53.1 in the first three months to between RMB~58.7 and RMB~90.8 in the last three months, whereas GLM-5.2, DeepSeek-V4-Flash, and Kimi K2.6 retain comparatively flat listing price trajectories.

\begin{figure}[!t]
\centering
\includegraphics[width=\columnwidth]{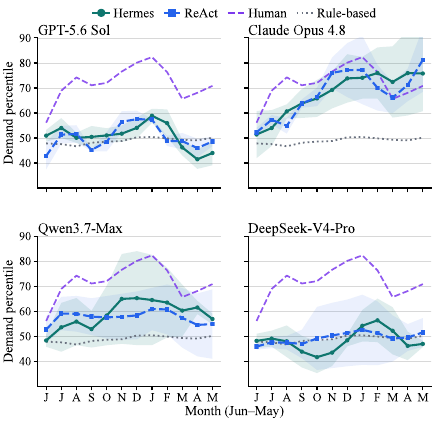}
\caption{Monthly Product Sourcing across four representative models.
Scores are listing-hour-weighted catalog demand percentiles of active products in each month.
Lines and bands show repeat means and standard deviations.}
\label{fig:monthly_product_sourcing_selected}
\end{figure}

Dynamic market demand makes Product Sourcing a continual portfolio allocation problem rather than a one-time selection decision.
Figure~\ref{fig:monthly_product_sourcing_selected} first measures monthly portfolio alignment, with Human rising from 56.1 in June to above 80 in December and January while Rule-based remains near the catalog median and LLM improvements are weaker or less consistent.
Figure~\ref{fig:operational_coherence_selected}(c) then reports realized profit, with Human peaking in winter while the winter profit gains of Hermes Qwen3.7-Max and GPT-5.6 Sol fade in spring.
Claude Opus 4.8 further shows that demand alignment alone is insufficient, since its stronger alignment in later months coincides with a shelf contraction from 37.3 to 12.0 products and no corresponding profit improvement.
Together, the figures show that effective long horizon operation requires alignment with changing demand, sufficient portfolio breadth, and conversion of that alignment into realized returns.

Memory traces further show how local errors become persistent policies.
In one Hermes Claude Opus 4.8 run, the agent falsely inferred that removing weak listings would concentrate traffic on the remaining products, while its shelf contracted from 47 active listings on Day 54 to three on Day 322 despite independent demand opportunities for every listing.
In one Hermes Qwen3.7-Max run, the agent misremembered Day 285 as the endpoint on Day 282 and stopped filling vacant slots with 83 days remaining, correcting the error only after simulated time advanced beyond the assumed endpoint.

\section{Conclusion}

We introduced MerchantBench to evaluate Long-Term Coherence through persistent seller-side e-commerce operation in a 365-day order-level environment grounded in 98,843 real e-commerce product records.
Across 48 runs involving eight LLMs and two agent frameworks, LLM agents exhibit a substantial gap from the Human baseline in system-level performance.
Trace analyses further show that weaker outcomes accompany declining operational activity, premature goal abandonment, and strategy changes that are not calibrated to accumulated evidence.
\bibliography{aaai2027}

\clearpage
\appendix
\section{Demand and Rating Dynamics}
\label{app:demand_and_rating_dynamics}

\paragraph{Listing Exposure.}
Let $a_{m,i,t}\geq 0$ denote the elapsed time in days since the current listing of product $i$ by merchant $m$ was activated.
We set $\ell_{m,i,t}=g(a_{m,i,t})$, where
\begin{equation}
g(a)=\left\{
\begin{array}{ll}
\ell_0+(1-\ell_0)a/T_r, & a\leq T_r,\\
\ell_{\min}+(1-\ell_{\min})e^{-\kappa(a-T_r)}, & a>T_r,
\end{array}
\right.
\label{eq:listing_exposure}
\end{equation}
This factor models a cold start through a linear exposure ramp followed by exponential decay toward $\ell_{\min}$.

\paragraph{Store Rating Dynamics.}
MerchantBench derives store reputation from realized terminal order outcomes and publishes the rating after each completed simulated day.
Each order $j$, except those resulting in Cancellation or insufficient balance failure, contributes an outcome-dependent experience score $u_j$ and evidence weight $v_j$.
For merchant $m$ on day $d$, the store rating is
\begin{equation}
R_{m,d}=\frac{\alpha R_0+
\sum_{j\in\mathcal{C}_{m,d}}\gamma^{d-1-d_j}v_j u_j}
{\alpha+\sum_{j\in\mathcal{C}_{m,d}}\gamma^{d-1-d_j}v_j}.
\label{eq:shop_rating}
\end{equation}
Here $\mathcal{C}_{m,d}$ contains the eligible orders resolved before day $d$, while $R_0$ and $\alpha$ define a prior that stabilizes sparse evidence and $\gamma$ discounts older outcomes.
The continuous rating is mapped to a discrete star level whose associated demand multiplier becomes $r_{m,t}$ in subsequent order generation.
The simulator separately aggregates the same outcome evidence for each listing as a diagnostic product rating, but listing ratings do not affect demand.

\section{Data Collection and Filtering}
\label{app:data_collection_and_filtering}

MerchantBench constructs its Product Catalog from real-world e-commerce data collected from 1688.
All source product and supplier attributes in the catalog originate from the platform.
Each Product Record contains a 365 day product level order history, and the data span ten first level product categories.
We remove records with missing product or supplier identifiers, missing product or supplier names, categories outside the selected ten categories, missing or nonpositive prices, invalid product or supplier attributes, or incomplete or invalid daily order histories.
After filtering, the resulting Product Catalog contains 98,843 Product Records from 36,576 suppliers.

Figure~\ref{fig:private_data_overview} summarizes the composition of the filtered data.
The catalog retains substantial variation across categories in both record coverage and supplier prices.
The joint risk distribution further shows that the calibrated data contain diverse combinations of Downstream Order Outcome probabilities and Upstream Supplier Event intensities.

Figure~\ref{fig:private_demand_patterns} presents the temporal structure of the 365 day demand histories.
The aggregate curve retains major shopping peaks and seasonal changes, while the representative product curves show distinct demand cycles for red envelope, electric fan, and hot water bag products.

\section{Full Operational Coherence Profiles}
\label{app:full_operational_coherence_profiles}

Figures~\ref{fig:operational_coherence_all_models} and~\ref{fig:operational_coherence_react} extend the monthly trajectory diagnostics to Human and all eight models under Hermes and ReAct, respectively.
Under Hermes, GPT-5.6 Sol remains almost fully active through February before declining to 81\% and 77\% in the final two months, while DeepSeek-V4-Flash stays between 70\% and 87\% throughout the year.
Qwen3.7-Max declines from 64\% to 31\%, Claude Opus 4.8 contracts from 46 to eight month end active listings, and Kimi K2.6 recovers to 78\% and 74\% effective windows only in the final two months.
Under ReAct, GPT-5.6 Sol sustains near complete activity, while Kimi K2.6 records the lowest Sustained Window Rate at 10.6\%.

\section{Net Asset Curves}
\label{app:net_asset_curves}

Figures~\ref{fig:net_assets_react} and~\ref{fig:net_assets_hermes} present the complete daily net asset curves under ReAct and Hermes, respectively.
Figure~\ref{fig:net_assets_all_models} compares the net asset curves of all eight models with the Human and Rule-based baselines.
Figure~\ref{fig:net_assets_by_run} shows the corresponding variation among individual runs.

\section{Scale and Unit Profitability}
\label{app:scale_and_unit_profitability}

Figure~\ref{fig:scale_and_unit_profitability} separates realized performance into order scale and net profit per order across all 48 LLM runs.
Qwen3.7-Max with Hermes produces the largest total profit in one run by combining high profit per order with moderate scale, whereas one Kimi K2.6 ReAct run reaches 3,004 orders at only RMB 11.1 net profit per order, illustrating that scale alone does not guarantee the highest return.
The wide dispersion across repeated runs shows that the evaluated configurations do not consistently reproduce the same balance between operating scale and unit profitability.

\par\medskip
\noindent\begin{minipage}{\columnwidth}
\centering
\includegraphics[width=\linewidth]{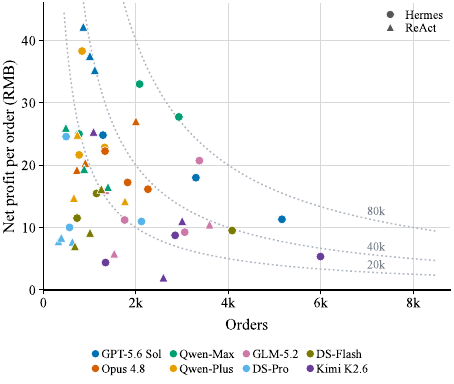}
\captionof{figure}{Order scale and unit profitability across 48 LLM runs.
Colors denote models and markers denote agent frameworks.
Dotted curves indicate equal cumulative net profit at RMB 20{,}000, 40{,}000, and 80{,}000.}
\label{fig:scale_and_unit_profitability}
\end{minipage}
\par\medskip

\section{Tool Use and Product Selection Analysis}
\label{app:tool_use_and_product_selection}

Across the 16 LLM configurations, greater tool use and broader product exploration are associated with higher final net assets, suggesting that sustained intervention matters in long horizon operation as shown in Figure~\ref{fig:efficiency_analysis}.
However, the dispersion around both fitted trends shows that activity volume alone is insufficient, since models differ in how effectively they translate actions into business value.

\par\medskip
\noindent\begin{minipage}{\columnwidth}
\centering
\includegraphics[width=\linewidth]{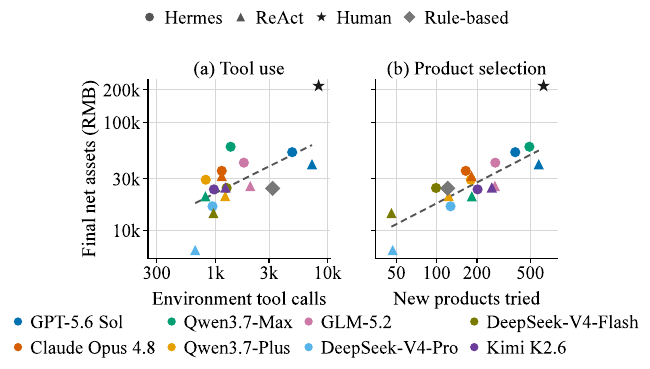}
\captionof{figure}{Final net assets against environment tool calls and new products tried.
Both axes are logarithmic, and dashed lines fit the 16 LLM configurations only.
Human and Rule-based are shown as references.}
\label{fig:efficiency_analysis}
\end{minipage}
\par\medskip

\section{Full Monthly Product Sourcing Analysis}
\label{app:full_monthly_product_sourcing}
\refstepcounter{figure}
\label{fig:monthly_product_sourcing_all_models}

For run $r$ and month $m$, let $q_{p,m}$ denote the demand percentile of product $p$ among the complete Product Catalog in that month, and let $w_{r,p,m}$ denote its active listing hours.
The Monthly Demand Alignment Percentile is
\begin{equation}
S_{r,m}
=
\frac{\sum_p w_{r,p,m}q_{p,m}}
{\sum_p w_{r,p,m}}.
\end{equation}
The catalog ranking uses real demand within each month, and the monthly denominator is the total active listing hours in that run and month.
A score of 80 means that an average listed product-hour belongs to a product with greater demand than 80\% of the catalog.

Figure~\ref{fig:monthly_product_sourcing_all_models} shows that Claude Opus 4.8 improves under both frameworks, while several other model and framework combinations plateau or decline during the year.
Human maintains the strongest demand alignment, whereas Rule-based remains close to the catalog median with little seasonal change.

\section{Time-aware Sourcing Gain}
\label{app:time_aware_sourcing_gain}

Monthly Demand Alignment can change even when an agent retains a fixed portfolio because product demand itself varies over time.
Time-aware Sourcing Gain therefore compares the actual monthly portfolio with a no-reallocation counterfactual constructed from the same run.
Let $W_{r,p}=\sum_m w_{r,p,m}$ denote the annual listing hours assigned to product $p$.
The no-reallocation counterfactual evaluates the same annual product mix in each month as
\begin{equation}
B_{r,m}
=
\frac{\sum_p W_{r,p}q_{p,m}}
{\sum_p W_{r,p}}.
\end{equation}
We define Time-aware Sourcing Gain as
\begin{equation}
G_r
=
\frac{\sum_m H_{r,m}\left(S_{r,m}-B_{r,m}\right)}
{\sum_m H_{r,m}},
\qquad
H_{r,m}=\sum_p w_{r,p,m}.
\end{equation}
Positive values indicate that the agent allocates listing exposure to products that better match the current month than its own fixed annual product mix.
An unchanged product mix yields a gain of zero.

Across the 48 LLM runs, higher Time-aware Sourcing Gain is positively associated with final net assets.
This result indicates that stronger seasonal portfolio reallocation accompanies better long horizon business outcomes, although final net assets also depend on pricing, cash flow, and order management decisions.

\par\medskip
\noindent\begin{minipage}{\columnwidth}
\centering
\includegraphics[width=\linewidth]{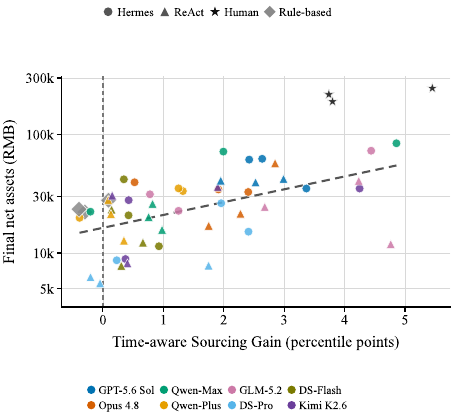}
\captionof{figure}{Time-aware Sourcing Gain and final net assets.
The gain compares actual monthly demand alignment with a counterfactual that holds each run's annual product mix fixed.
The dashed line fits the 48 LLM runs, with Human and Rule-based shown as references.}
\label{fig:time_aware_sourcing_gain}
\end{minipage}
\par\medskip

\section{Hermes Case Study}
\label{app:hermes_case_study}

This section examines how the evaluated models use the additional Hermes capabilities listed in Tables~\ref{tab:hermes_tool_inventory} and~\ref{tab:hermes_skill_inventory}, focusing on programmatic code use and on the creation and revision of skills.

\subsection{Code Use}

Native code use differs sharply across the 24 Hermes runs.
GPT-5.6 Sol dominates native code use, issuing 15 \texttt{execute\_code} calls across two runs and 13 in one run.
At the first decision window it converts 50 candidate products into a priced launch portfolio by applying $p=\max(1.70c,c+6)$ to each procurement cost $c$ and rounding the result upward to a price ending in 0.9, then passes the computed pairs to MerchantBench listing calls.
A companion script validates the portfolio by printing the product count, average cost, minimum margin, and total procurement outlay before any listing action.
Claude Opus 4.8, Qwen3.7-Max, and DeepSeek-V4-Pro each invoke \texttt{execute\_code} once, while Qwen3.7-Max and DeepSeek-V4-Flash each invoke \texttt{terminal} once.
GLM-5.2, Qwen3.7-Plus, and Kimi K2.6 never invoke either code tool.
Claude Opus 4.8 instead channels its analysis through 261 \texttt{memory} calls across its three runs, maintaining day stamped operating hypotheses labeled as validated or rejected.
Hermes code capabilities therefore amplify models that already favor quantitative bookkeeping while leaving purely verbal operators unchanged.

\subsection{Skill Evolution}

Hermes periodically runs a background review that can distill the accumulated trace into a named skill and later patch it.
Final Hermes profiles contain a run created RealShop skill in 17 of 24 runs and for seven of the eight models.
Seventeen of the 18 created skills record at least one subsequent use, and the only unused skill was created by Qwen3.7-Max.
GPT-5.6 Sol, Kimi K2.6, and Qwen3.7-Plus create a skill in all three runs, while Claude Opus 4.8 creates none.
GPT-5.6 Sol creates a \texttt{realshop-store-operations} skill in each run, with 9, 15, and 7 patches and 17, 10, and 13 recorded uses.
These skills frame the task as repeated portfolio allocation and prescribe per window procedures for evidence collection, risk handling, unit economics, portfolio revision, and verification.
DeepSeek-V4-Pro and DeepSeek-V4-Flash each reach 20 patches in one run, while Qwen3.7-Plus creates two complementary store operation and optimization skills in one run.
These traces show that Hermes can convert experience into explicit procedural knowledge, but the quality of the resulting skills ranges from evidence cited rule revision to unstructured accumulation and duplication.

\section{Detailed Experimental Configuration}
\label{app:detailed_experimental_configuration}

\paragraph{Evaluation Protocol.}
The Rule-based baseline completes three runs.
It handles abnormal states through daily checks.
It delists products with no sales for seven consecutive days or products affected by Price Change, Product Delisting, or Shipment Delay, then fills available listing slots with new products selected using keywords from the daily market report.
Three participants with no prior e-commerce operating experience each complete one run spanning 365 simulated days through the human operations dashboard over five calendar days.
We report the mean across three LLM or Rule-based runs or across the three human participants.

\paragraph{Store Configuration.}
Each agent operates a store initialized with a cash balance of RMB 2,000 and a security deposit of RMB 1,000 and supports at most 50 active listings.
The fixed fines are RMB 8 for Return and Refund, RMB 5 for Bad Review, Stockout, or insufficient balance, and RMB 3 for Late Shipment.
Cancellation and Returnless Refund incur no additional fine.
These fine settings are consistent with the corresponding real-world platform rules.

\paragraph{Context Management.}
To support interaction over the full 365 day horizon, both frameworks compress long interaction histories.
When a ReAct history reaches 160,000 tokens, the evaluated model receives a reminder to summarize important information into persistent memory before the history is truncated to the most recent 30,000 tokens.
Hermes uses its default context summarization procedure and sets the summary model to the same evaluated model.

\paragraph{Evaluation Metrics.}
Net Profit Margin is terminal net profit divided by GMV.
Average Store Rating and Average Active Listings are daily means of the published store rating and active listing count.
Order Anomaly Rate is the share of generated orders affected by at least one cancellation, stockout, insufficient balance failure, late shipment, refund, or bad review.
Sustained Window Rate is the minimum share of scheduled decision windows containing at least one environment tool call across all rolling 30 day periods.
Total Tool Calls excludes calls that end a decision window and all framework internal tools.

\paragraph{Demand and Supplier Configuration.}
Listing exposure uses $\ell_0=0.2$, $T_r=14$ days, $\kappa=0.0092$, and $\ell_{\min}=0.10$.
Supplier inventory is initialized at 20 to 399 units, with capacities of 50 to 499 units and hourly replenishment of 1 to 19 units.
Baseline supplier dispatch and logistics times span 1 to 47 and 12 to 72 hours.
Supplier abnormalities recover after 168 to 672 hours, Price Change applies a factor sampled from 0.9 to 1.5, and Shipment Delay adds 12 to 96 hours.

\paragraph{Order Lifecycle Configuration.}
The default promised shipment time is 48 hours.
The transition from Delivered to Settled and the realization of outcomes after delivery are each sampled within 168 hours.

\paragraph{Rating Configuration.}
The outcome scores $u_j$ are $4.5$, $3.0$, $2.0$, $1.5$, $1.0$, and $1.0$, and the evidence weights $v_j$ are $1.0$, $1.0$, $1.0$, $2.0$, $2.0$, and $3.0$, ordered as a normal transition to Settled, Late Shipment, Return and Refund, Returnless Refund, Bad Review, and Stockout.
Store ratings use $R_0=4.0$, $\alpha=20$, and $\gamma=2^{-1/30}$.
The rating thresholds $2.50$, $3.30$, $3.80$, and $4.20$ map to demand multipliers $0.10$, $0.35$, $0.80$, $1.00$, and $1.20$ across the five resulting intervals.
Diagnostic listing ratings use the same initial rating and prior weight with a 90 day evidence half life.

\section{Tool Sets and Skills}
\label{app:tool_set}

ReAct uses only the 26 MerchantBench tools through which the agent accesses observable fields and controls the store.
Table~\ref{tab:merchant_tool_inventory} lists the complete MerchantBench tool set.
Among them, the \texttt{get\_daily\_report} tool returns the report published for the current simulation date and states that its evidence is current only through the previous date.
Table~\ref{tab:daily_market_report_example} gives an English translation of the Daily Market Report for June 10, 2025.
Hermes provides the same MerchantBench tools together with the built in tools and skills listed in Tables~\ref{tab:hermes_tool_inventory} and~\ref{tab:hermes_skill_inventory} \cite{hermesagent2026}.

\section{Agent Inputs and Observable Fields}
\label{app:agent_inputs_and_observable_fields}

The environment gives each evaluated agent one system prompt at registration and a compact observation at every 12 hour decision window.
Both frameworks receive the shared MerchantBench task prompt in Table~\ref{tab:react_system_prompt} and access the same 26 MerchantBench tools.
For ReAct, the MerchantBench task prompt is the complete system prompt and the 26 tools are the entire action space.
Hermes wraps the same task prompt within its official runtime template, shown in Table~\ref{tab:hermes_system_prompt}, and augments the 26 MerchantBench tools with its built in tools and skills.
Table~\ref{tab:observation_example} gives a representative observation containing order transitions, an upstream price change, current finances, and the shop rating.

MerchantBench is partially observable because the merchant interface returns current public and realized operational evidence while retaining future demand, event hazards, and presampled outcomes inside the simulator.
Tables~\ref{tab:product_field_visibility} to~\ref{tab:order_field_visibility} summarize this boundary.
Visible fields are returned directly by at least one merchant tool.
Some visible order fields remain empty until the corresponding lifecycle event is realized.
Hidden fields are never returned through merchant tools.

\clearpage

\begin{figure*}[p]
\centering
\includegraphics[width=0.98\textwidth]{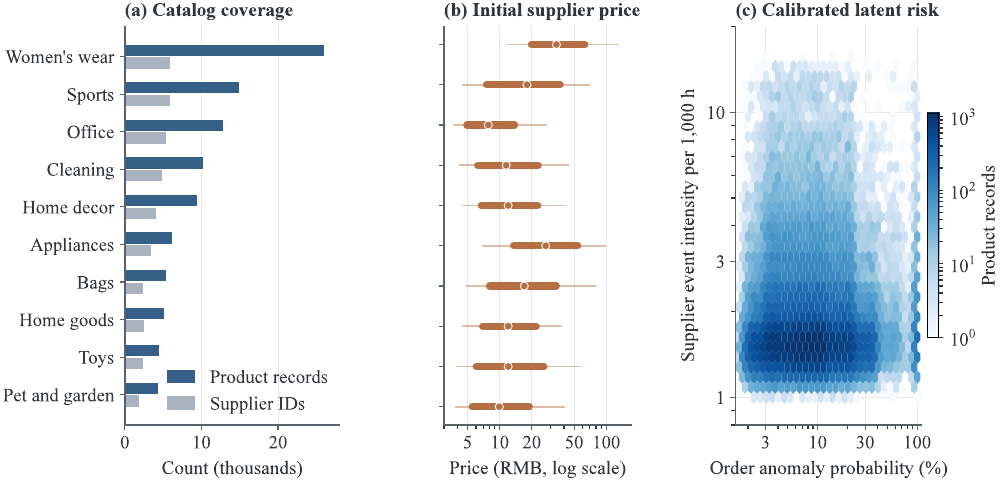}
\caption{Composition and calibrated distributions of the filtered data.
Panel (a) reports the numbers of Product Records and unique supplier IDs within each category.
Panel (b) shows procurement price distributions within each category.
Panel (c) shows the joint distribution of Downstream Order Outcome probability and Upstream Supplier Event intensity.}
\label{fig:private_data_overview}
\end{figure*}

\begin{figure*}[p]
\centering
\includegraphics[width=0.98\textwidth]{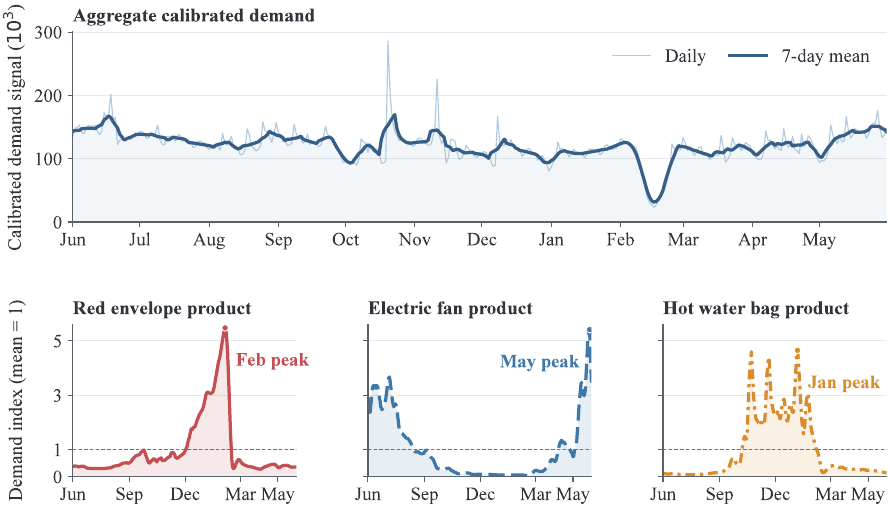}
\caption{Temporal demand patterns in the filtered data.
The top panel presents aggregate daily demand and its seven day mean.
The bottom panels show normalized demand histories for representative red envelope, electric fan, and hot water bag Product Records.}
\label{fig:private_demand_patterns}
\end{figure*}

\begin{figure*}[p]
\centering
\includegraphics[width=0.98\textwidth]{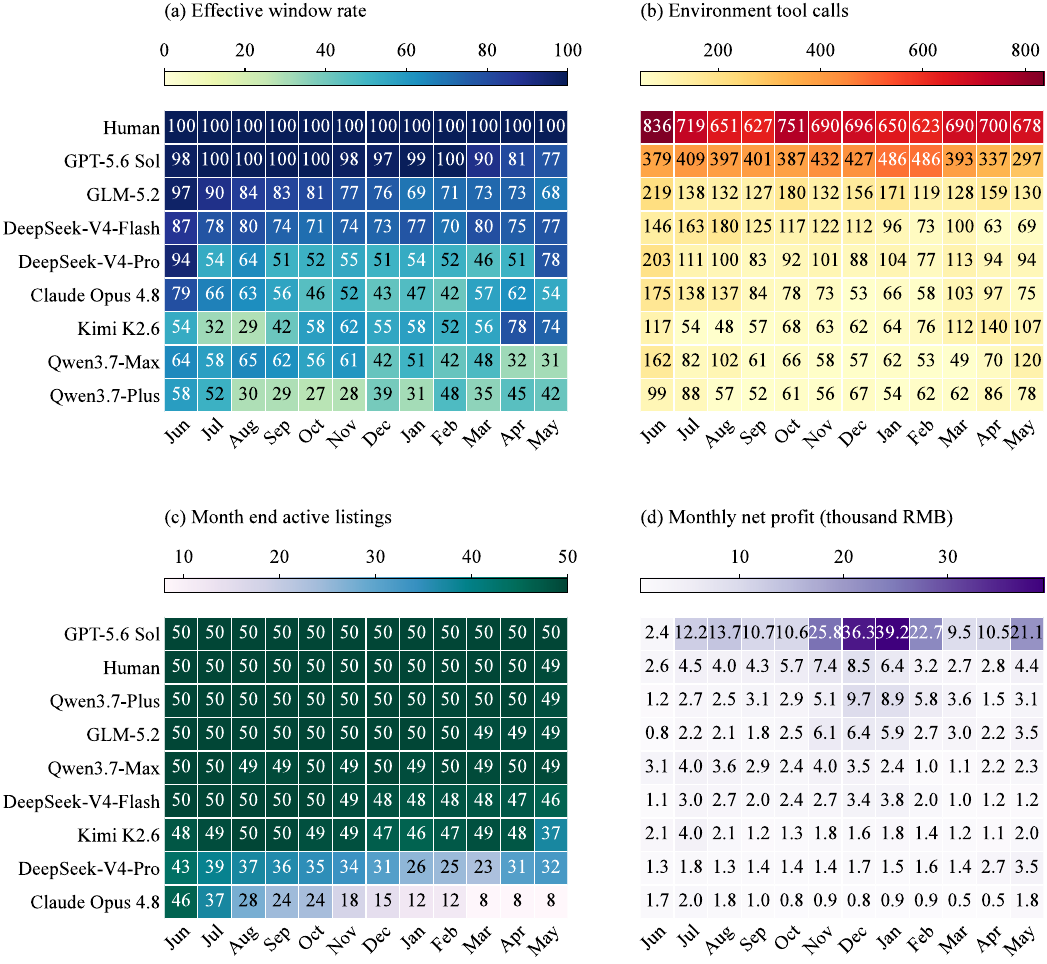}
\caption{Monthly Operational Coherence profiles for the Human baseline and each of the eight Hermes models.
Panels (a), (b), (c), and (d) report effective window rate, environment tool calls, month end active listings, and monthly net profit.}
\label{fig:operational_coherence_all_models}
\end{figure*}

\begin{figure*}[p]
\centering
\includegraphics[width=0.98\textwidth]{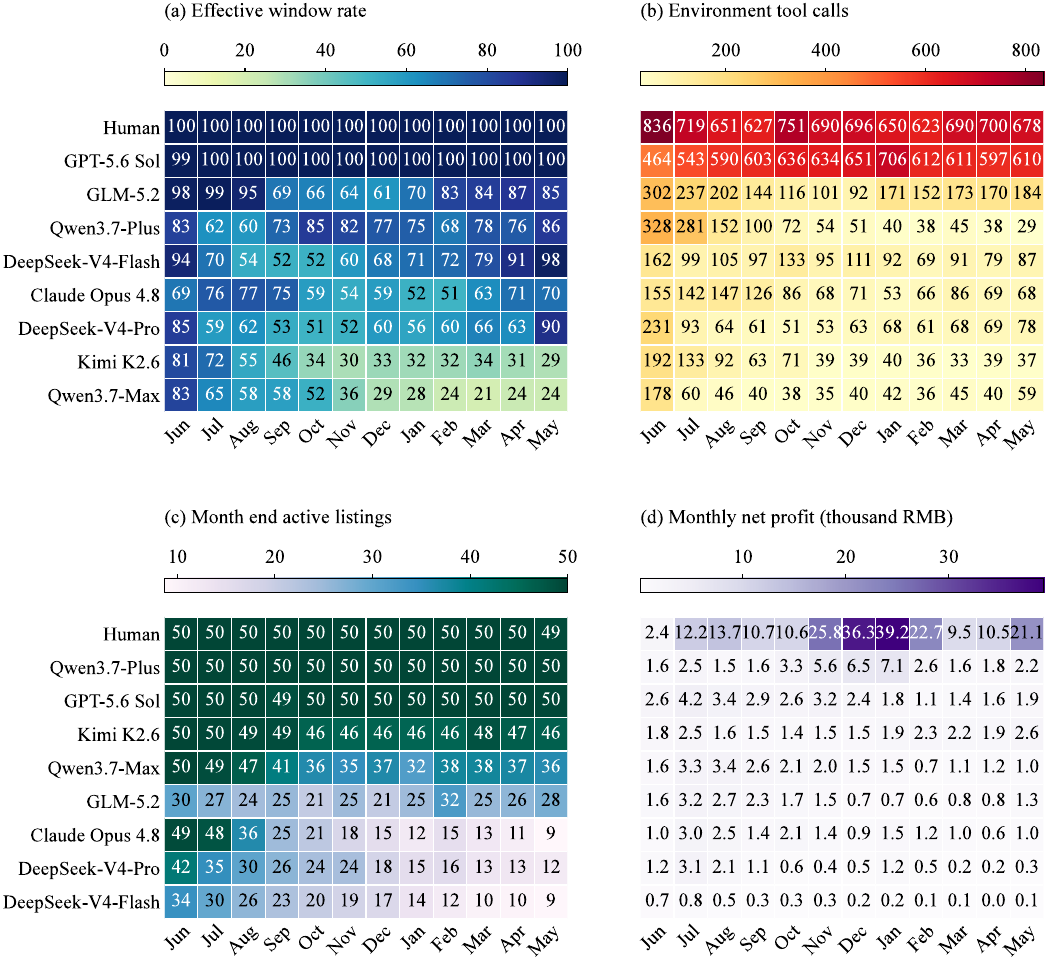}
\caption{Monthly Operational Coherence profiles for the Human baseline and each of the eight ReAct models.
Panels (a), (b), (c), and (d) report effective window rate, environment tool calls, month end active listings, and monthly net profit.}
\label{fig:operational_coherence_react}
\end{figure*}

\begin{figure*}[p]
\centering
\includegraphics[width=0.98\textwidth]{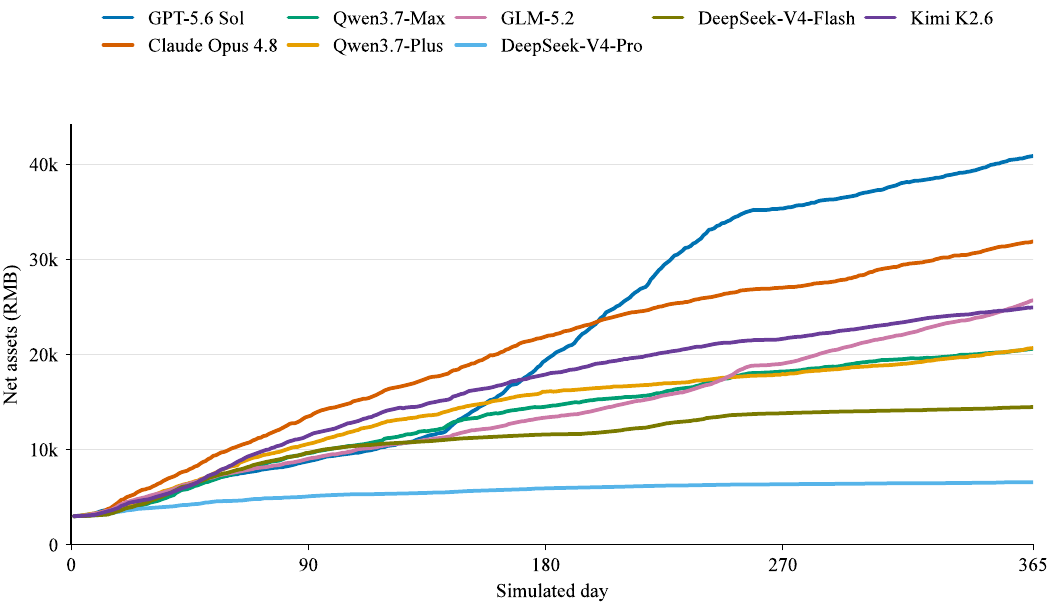}
\caption{Daily net asset curves for all eight models under ReAct over 365 simulated days.}
\label{fig:net_assets_react}
\end{figure*}

\begin{figure*}[p]
\centering
\includegraphics[width=0.98\textwidth]{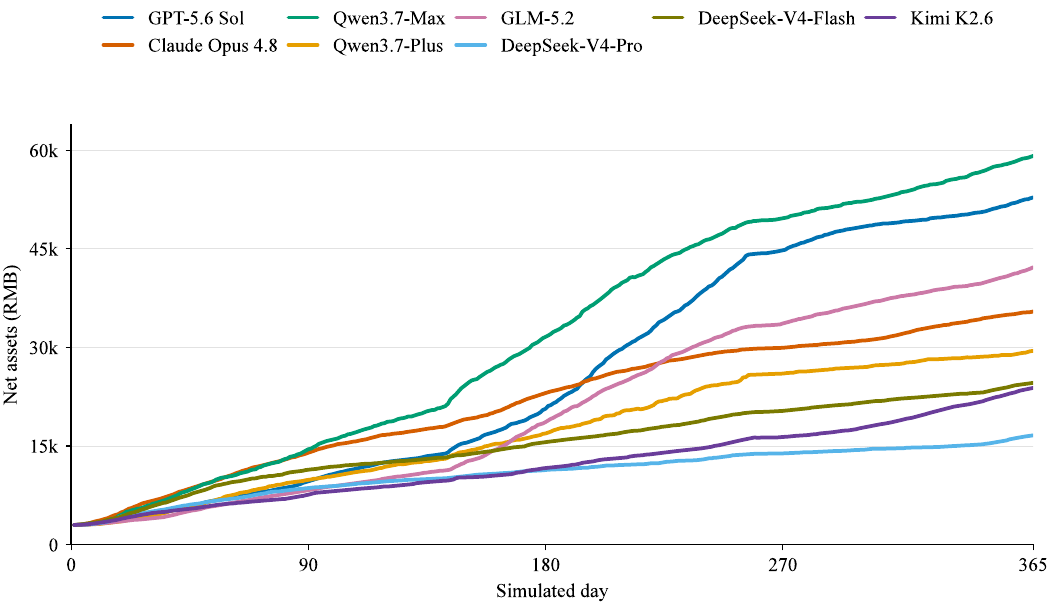}
\caption{Daily net asset curves for all eight models under Hermes over 365 simulated days.}
\label{fig:net_assets_hermes}
\end{figure*}

\begin{figure*}[p]
\centering
\includegraphics[width=0.98\textwidth]{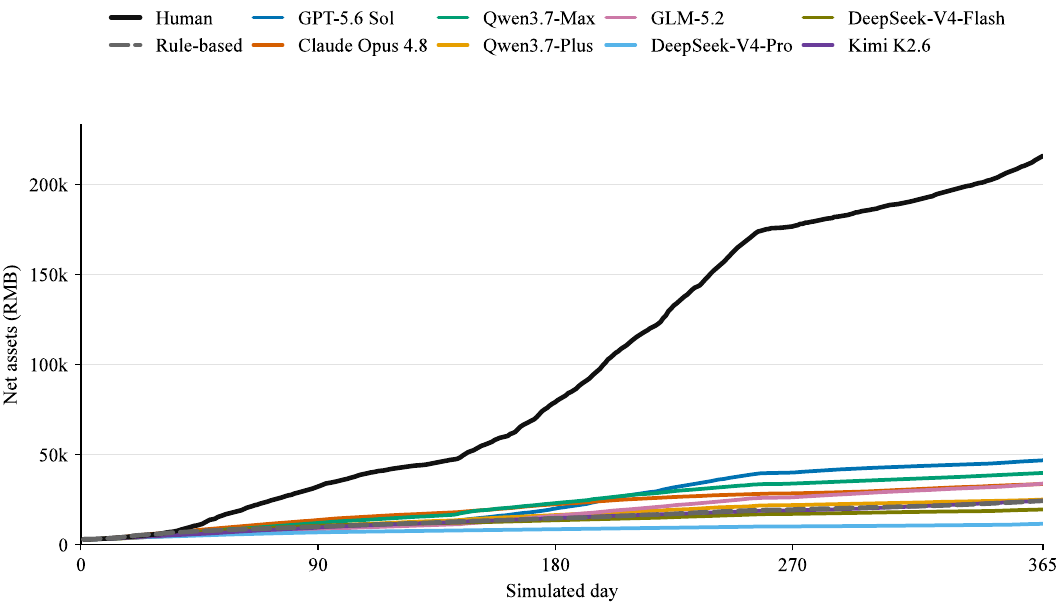}
\caption{Daily net asset curves for the Human and Rule-based baselines and all eight models over 365 simulated days.}
\label{fig:net_assets_all_models}
\end{figure*}

\begin{figure*}[p]
\centering
\includegraphics[width=0.97\textwidth]{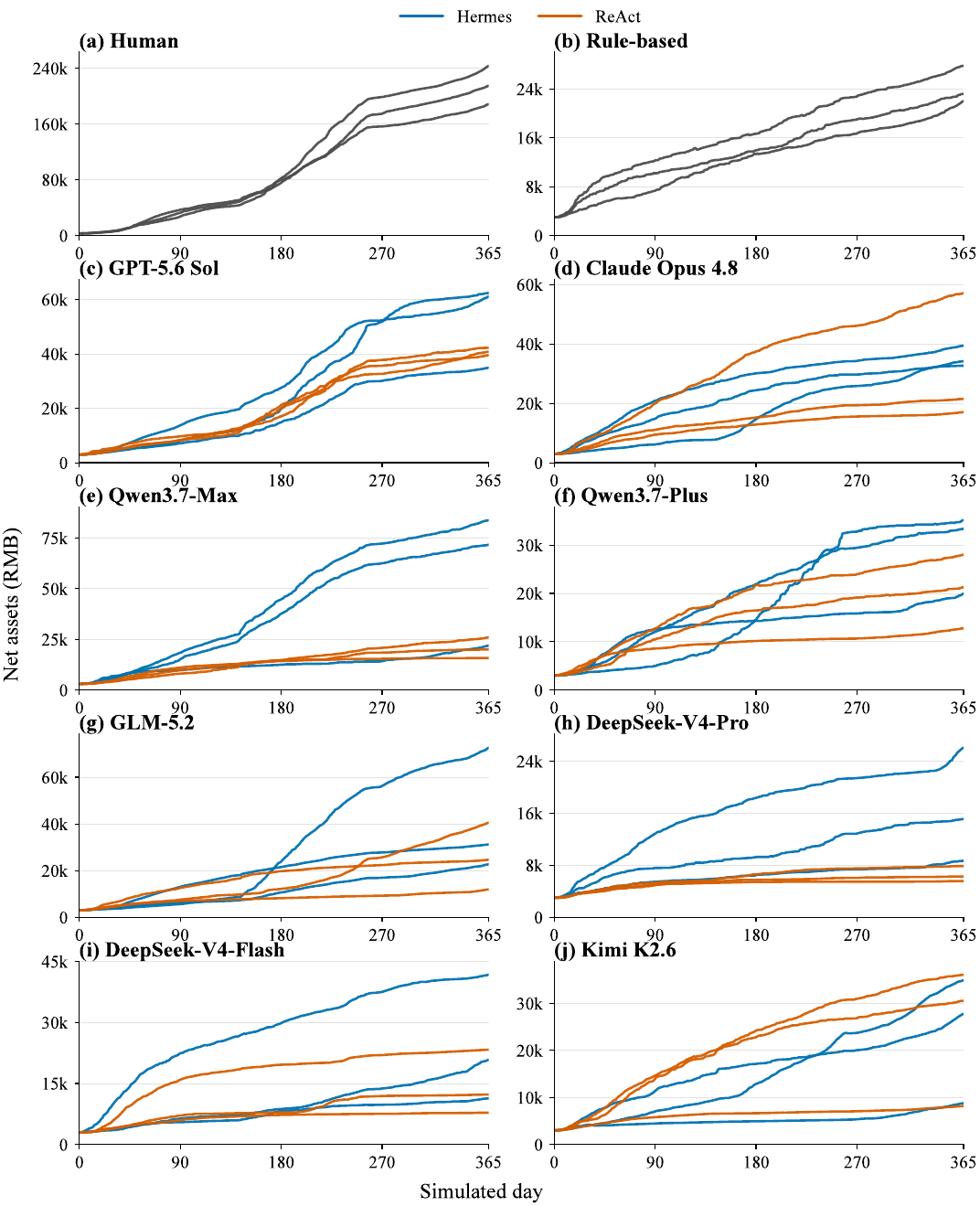}
\caption{Daily net asset curves for all individual runs.
The first row shows the Human and Rule-based baselines, and the remaining panels show the eight models.
Hermes and ReAct runs appear together within each model panel, and vertical axis ranges are set independently across panels.}
\label{fig:net_assets_by_run}
\end{figure*}

\begin{figure*}[p]
\centering
\includegraphics[width=0.98\textwidth]{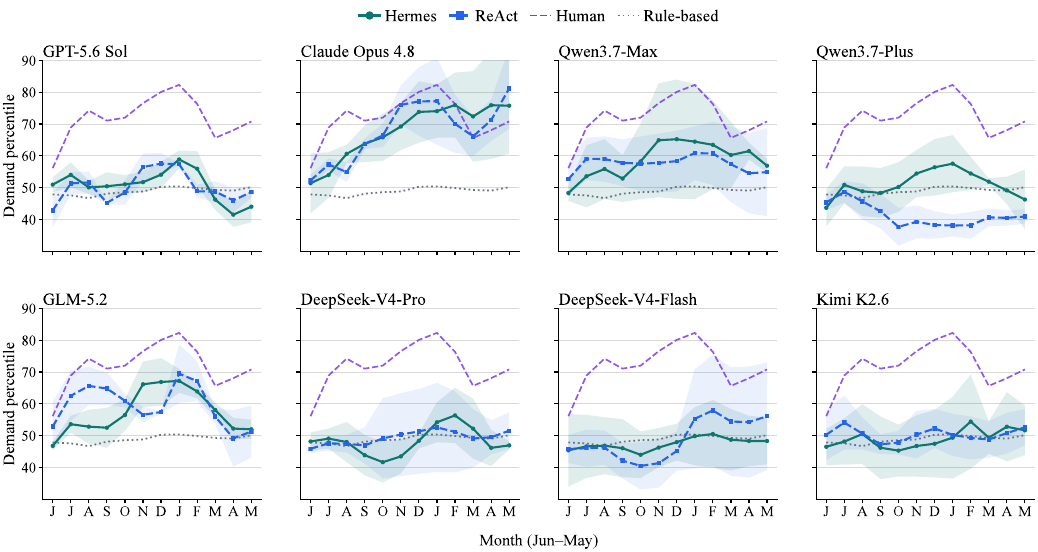}
\caption*{\textbf{Figure~\ref{fig:monthly_product_sourcing_all_models}:} Monthly Product Sourcing across all eight models.
Monthly Demand Alignment Percentile is the listing-hour-weighted mean percentile of active products after ranking the complete catalog by that month's real demand.
Lines and bands show repeat means and standard deviations, with Human and Rule-based as shared references.}
\end{figure*}

\begin{table*}[p]
\centering
{\small
\renewcommand{\arraystretch}{0.94}
\begin{tabular*}{\textwidth}{@{\extracolsep{\fill}}p{0.255\textwidth}cp{0.60\textwidth}@{}}
\toprule
\textbf{Tool} & \textbf{Access} & \textbf{Description} \\
\midrule
\multicolumn{3}{c}{\textbf{Product Sourcing}} \\
\midrule
get\_\allowbreak daily\_\allowbreak report & Read & Returns the daily market report for the current simulation date with market news and opportunity signals \\
search\_\allowbreak products & Read & Searches the visible Product Catalog using public fields \\
get\_\allowbreak product\_\allowbreak detail & Read & Returns visible product, logistics, rating, and supplier fields \\
get\_\allowbreak supplier\_\allowbreak profile & Read & Returns the public supplier profile and visible product count \\
list\_\allowbreak supplier\_\allowbreak products & Read & Lists the currently visible products from one supplier \\
\midrule
\multicolumn{3}{c}{\textbf{Listing and Pricing Control}} \\
\midrule
list\_\allowbreak product & Write & Adds products to the store at specified selling prices \\
delist\_\allowbreak product & Write & Removes products from the store \\
adjust\_\allowbreak price & Write & Changes selling prices for active listings \\
review\_\allowbreak my\_\allowbreak listings & Read & Reviews listing age, sales velocity, fines, and fulfillment backlog \\
query\_\allowbreak my\_\allowbreak listings & Read & Returns current listings with cumulative sales, profit, and fines \\
query\_\allowbreak store\_\allowbreak performance & Read & Summarizes store outcomes by day or week \\
query\_\allowbreak product\_\allowbreak sales\_\allowbreak stats & Read & Ranks product outcomes and reports abnormality counts \\
\midrule
\multicolumn{3}{c}{\textbf{Cash-Flow Management}} \\
\midrule
query\_\allowbreak balance & Read & Returns the cash balance, security deposit, funds in transit, receivables, and fines \\
get\_\allowbreak store\_\allowbreak snapshot & Read & Summarizes orders, supply, cash, listings, and store rating \\
query\_\allowbreak platform\_\allowbreak rules & Read & Returns capital, settlement, penalty, and closure rules \\
query\_\allowbreak cash\_\allowbreak pipeline & Read & Summarizes receivable aging and active order cost exposure \\
\midrule
\multicolumn{3}{c}{\textbf{Supplier and Order Monitoring}} \\
\midrule
query\_\allowbreak supply\_\allowbreak chain\_\allowbreak anomalies & Read & Returns new or current supplier abnormalities and affected listings \\
query\_\allowbreak my\_\allowbreak orders & Read & Searches historical orders with logistics and accounting fields \\
query\_\allowbreak open\_\allowbreak orders & Read & Returns active orders with fulfillment timing and economics \\
query\_\allowbreak order\_\allowbreak updates & Read & Returns status changes since the previous observation window \\
query\_\allowbreak order\_\allowbreak detail & Read & Returns one order's full status timeline, accounting, and penalties \\
\midrule
\multicolumn{3}{c}{\textbf{Agent Support and Control}} \\
\midrule
read\_\allowbreak memory\_\allowbreak doc & Read & Reads the run local agent memory document \\
write\_\allowbreak memory\_\allowbreak doc & Write & Replaces the run local agent memory document \\
get\_\allowbreak observation & Read & Returns the current rendered observation \\
list\_\allowbreak tools & Read & Returns tool schemas after scenario filtering \\
end\_\allowbreak of\_\allowbreak step & Control & Releases the current decision window \\
\bottomrule
\end{tabular*}
}
\caption{MerchantBench merchant tool inventory.}
\label{tab:merchant_tool_inventory}
\end{table*}

\begin{table*}[p]
\centering
{\setlength{\fboxsep}{7pt}
\setlength{\fboxrule}{0.45pt}
\begin{minipage}{0.945\textwidth}
\vspace{2pt}
\begin{lstlisting}[style=artifactcontent]
June 10 Market Opportunity Brief

Report date: June 10, 2025
Data current through: June 9, 2025

Yesterday's news

- The State Administration for Market Regulation issued compliance guidance for the June 18 promotion across general, livestream, and cross-border e-commerce platforms. The guidance prohibits discriminatory pricing, false marketing, fabricated transactions, and subsidy fraud, and asks platforms to strengthen monitoring of livestreams and hosts.
- Consumer subsidies expanded from eight to twelve appliance categories. Eligible appliances receive a 20 percent subsidy capped at RMB 2,000, while selected digital products receive a 15 percent subsidy capped at RMB 500. Subsidy claims on Taobao and Tmall during the first four hours tripled relative to the previous Double 11 promotion, and first-day sales on Meituan Instashopping increased by 200 percent year over year.
- The June 18 promotion period lengthened. Tmall runs presales from May 13 through June 20, Xiaohongshu launched a thirty-day marketplace promotion on June 1, and Meituan entered the promotion through instant retail. Competition now spans assortment, fulfillment, and shopping scenarios rather than price alone.

Overall trend

Search volume reached 19.052 million yesterday, up 9.08 percent from seven days earlier despite a daily pullback. The top 100 terms covered 31.7 percent of searches across 1,813 keywords. Appliances rose 45.8 percent, other goods rose 5.8 percent, and summer apparel and womenswear rose 9.3 percent. Cleaning, sports and outdoor goods, and toys declined by 10.4, 8.7, and 19.3 percent. Traffic concentrated on cooling appliances, air conditioners, swimwear, and Father's Day products.

Category dynamics

- Appliances. Neck fans rose 141.6 percent, desktop fans rose 133.3 percent, electric fans rose 118.8 percent, and Xiaomi air conditioners rose 112.1 percent. Small fans and electric fans both entered the top ten.
- Other goods. Father's Day gifts entered at rank 65 with 257.9 percent growth. Ice makers remained strong with 65.2 percent growth, and classmate albums rose 44.5 percent.
- Summer apparel and womenswear. Women's summer shorts rose 54.2 percent, women's shorts rose 45.2 percent, and women's swimwear newly entered with 44.7 percent growth.
- Student and office supplies. A4 printing paper newly entered with 28.2 percent growth. Notebooks and correction tape rose 10.7 and 7.1 percent.
- Home and cleaning goods. Disposable bath towels newly entered at rank 93 with 7.0 percent growth, indicating the start of a summer expansion window.

Anomalous search signals

- #40 Gree air conditioner. 54,687 searches, up 93.9 percent from rank 118.
- #55 neck fan. 48,005 searches, up 141.6 percent from rank 205.
- #59 Midea air conditioner. 46,505 searches, up 67.9 percent from rank 123.
- #65 Father's Day gift. 43,519 searches, up 257.9 percent from rank 402.
\end{lstlisting}
\vspace{2pt}
\end{minipage}
}
\caption{Daily market report for June 10, 2025, shown here in English translation.}
\label{tab:daily_market_report_example}
\end{table*}

\begin{table*}[p]
\ContinuedFloat
\centering
{\setlength{\fboxsep}{7pt}
\setlength{\fboxrule}{0.45pt}
\begin{minipage}{0.945\textwidth}
\vspace{2pt}
\begin{lstlisting}[style=artifactcontent]
- #76 A4 printing paper. 39,238 searches, up 28.2 percent from rank 103.
- #78 ice maker. 39,148 searches, up 65.2 percent from rank 149.
- #86 pedestal fan. 36,888 searches, up 76.4 percent from rank 183.
- #97 women's swimwear. 32,817 searches, up 44.7 percent from rank 159.
- #98 desktop fan. 32,806 searches, up 133.3 percent from rank 320.
- #23 Xiaomi air conditioner. 70,369 searches, up 112.1 percent and 65 ranks.

News-linked suggestions

- Father's Day gifts show an immediate sales window before June 15. Candidate products include electric shavers, belts, wallets, men's personal care products, and massage devices.
- Merchants using subsidized prices should retain invoices and trade-in documentation to reduce compliance risk.
- A 15 percent direct discount has become common on Douyin and Xiaohongshu. Highly ranked products may need a lower markup to remain competitive during the promotion.
\end{lstlisting}
\vspace{2pt}
\end{minipage}
}
\caption{Daily market report for June 10, 2025, with the English translation continued.}
\end{table*}

\begin{table*}[p]
\centering
\footnotesize
\renewcommand{\arraystretch}{1.02}
\begin{tabular*}{\textwidth}{@{\extracolsep{\fill}}p{0.24\textwidth}p{0.10\textwidth}p{0.60\textwidth}@{}}
\toprule
\textbf{Tool} & \textbf{Access} & \textbf{Description} \\
\midrule
\multicolumn{3}{c}{\textbf{Execution and Files}} \\
\midrule
\texttt{terminal} & Execute & Executes shell commands in a persistent environment \\
\texttt{process} & Manage & Monitors and controls background processes \\
\texttt{execute\_code} & Execute & Runs Python programs that call Hermes tools and process their outputs \\
\texttt{read\_file} & Read & Reads text files with line numbers and pagination \\
\texttt{write\_file} & Write & Creates or replaces files and checks supported formats \\
\texttt{patch} & Write & Applies targeted file edits and returns a unified diff \\
\texttt{search\_files} & Read & Searches file names and contents \\
\midrule
\multicolumn{3}{c}{\textbf{Memory and Skills}} \\
\midrule
\texttt{memory} & Write & Stores durable facts that persist across sessions \\
\texttt{session\_search} & Read & Searches messages from previous Hermes sessions \\
\texttt{skills\_list} & Read & Lists available skills and their descriptions \\
\texttt{skill\_view} & Read & Loads skill instructions and linked resources \\
\texttt{skill\_manage} & Write & Creates, revises, or deletes skills \\
\midrule
\multicolumn{3}{c}{\textbf{Planning and Coordination}} \\
\midrule
\texttt{todo} & Manage & Maintains the task list for the current session \\
\texttt{clarify} & Interact & Requests clarification, feedback, or a decision from the user \\
\texttt{delegate\_task} & Delegate & Assigns independent tasks to subagents \\
\midrule
\multicolumn{3}{c}{\textbf{Projects and Output}} \\
\midrule
\texttt{project\_list} & Read & Lists available project workspaces \\
\texttt{project\_create} & Write & Creates and activates a project workspace \\
\texttt{project\_switch} & Write & Switches the active project workspace \\
\texttt{text\_to\_speech} & Generate & Converts text into speech audio \\
\texttt{image\_generate} & Generate & Generates or edits images from prompts and references \\
\bottomrule
\end{tabular*}
\caption{Built in Hermes tools provided by the official architecture.}
\label{tab:hermes_tool_inventory}
\end{table*}

\begin{table*}[p]
\centering
\footnotesize
\renewcommand{\arraystretch}{1.02}
\begin{tabular*}{\textwidth}{@{\extracolsep{\fill}}p{0.25\textwidth}p{0.16\textwidth}p{0.53\textwidth}@{}}
\toprule
\textbf{Skill} & \textbf{Category} & \textbf{Description} \\
\midrule
apple-notes & Apple & Creates, searches, and edits Apple Notes \\
apple-reminders & Apple & Adds, lists, and completes Apple Reminders \\
findmy & Apple & Tracks Apple devices and AirTags \\
imessage & Apple & Sends and receives iMessages and SMS \\
claude-code & Autonomous agents & Delegates coding tasks to Claude Code \\
codex & Autonomous agents & Delegates coding tasks to OpenAI Codex \\
hermes-agent & Autonomous agents & Configures and extends the Hermes Agent codebase \\
opencode & Autonomous agents & Delegates coding and review tasks to OpenCode \\
computer-use & General & Operates desktop interfaces through visual interaction \\
architecture-diagram & Creative & Creates architecture and infrastructure diagrams \\
ascii-art & Creative & Generates and transforms ASCII art \\
ascii-video & Creative & Converts video and audio into ASCII video \\
baoyu-infographic & Creative & Produces infographics using reusable layouts and styles \\
claude-design & Creative & Designs standalone HTML artifacts \\
comfyui & Creative & Generates images, video, and audio with ComfyUI \\
design-md & Creative & Authors and validates DESIGN.md specifications \\
excalidraw & Creative & Creates hand drawn Excalidraw diagrams \\
humanizer & Creative & Revises text to remove formulaic AI phrasing \\
manim-video & Creative & Produces mathematical and algorithmic animations \\
p5js & Creative & Creates interactive p5.js sketches and generative art \\
popular-web-designs & Creative & Applies established web interface design systems \\
pretext & Creative & Supports interactive creative browser demonstrations \\
sketch & Creative & Produces alternative HTML interface mockups \\
songwriting-and-ai-music & Creative & Supports songwriting and AI music prompting \\
touchdesigner-mcp & Creative & Controls TouchDesigner through an MCP interface \\
\bottomrule
\end{tabular*}
\caption{Built in Hermes skills and their functions.}
\label{tab:hermes_skill_inventory}
\end{table*}

\begin{table*}[p]
\ContinuedFloat
\centering
\footnotesize
\renewcommand{\arraystretch}{1.02}
\begin{tabular*}{\textwidth}{@{\extracolsep{\fill}}p{0.25\textwidth}p{0.16\textwidth}p{0.53\textwidth}@{}}
\toprule
\textbf{Skill} & \textbf{Category} & \textbf{Description} \\
\midrule
jupyter-live-kernel & Data science & Performs iterative analysis in a persistent Jupyter kernel \\
dogfood & General & Conducts exploratory testing of web applications \\
himalaya & Email & Manages email through the Himalaya command line interface \\
codebase-inspection & GitHub & Measures codebase size, languages, and composition \\
github-auth & GitHub & Configures tokens, keys, and command line authentication \\
github-code-review & GitHub & Reviews pull request diffs and inline comments \\
github-issues & GitHub & Creates and manages GitHub issues \\
github-pr-workflow & GitHub & Manages branches, commits, checks, and pull requests \\
github-repo-management & GitHub & Clones, creates, forks, and maintains repositories \\
gif-search & Media & Searches and downloads GIF content \\
heartmula & Media & Generates songs from lyrics and style tags \\
songsee & Media & Extracts and visualizes audio features \\
youtube-content & Media & Converts YouTube transcripts into written content \\
huggingface-hub & MLOps & Searches, downloads, and uploads models and datasets \\
evaluating-llms-harness & MLOps & Evaluates language models with standard benchmarks \\
weights-and-biases & MLOps & Tracks experiments, sweeps, and model artifacts \\
llama-cpp & MLOps & Runs local GGUF model inference \\
serving-llms-vllm & MLOps & Serves language models with vLLM \\
audiocraft-audio-generation & MLOps & Generates music and sound with AudioCraft \\
segment-anything-model & MLOps & Performs prompt based image segmentation \\
obsidian & Note taking & Reads, searches, creates, and edits Obsidian notes \\
\bottomrule
\end{tabular*}
\caption{Built in Hermes skills and their functions, continued.}
\end{table*}

\begin{table*}[p]
\ContinuedFloat
\centering
\footnotesize
\renewcommand{\arraystretch}{1.02}
\begin{tabular*}{\textwidth}{@{\extracolsep{\fill}}p{0.25\textwidth}p{0.16\textwidth}p{0.53\textwidth}@{}}
\toprule
\textbf{Skill} & \textbf{Category} & \textbf{Description} \\
\midrule
airtable & Productivity & Manages Airtable records and queries \\
google-workspace & Productivity & Operates Gmail, Calendar, Drive, Docs, and Sheets \\
maps & Productivity & Provides geocoding, points of interest, routes, and time zones \\
nano-pdf & Productivity & Edits PDF text and document metadata \\
notion & Productivity & Manages Notion pages and databases \\
ocr-and-documents & Productivity & Extracts text from PDFs and scanned documents \\
petdex & Productivity & Installs and selects animated Hermes mascots \\
powerpoint & Productivity & Creates and edits presentation decks \\
teams-meeting-pipeline & Productivity & Operates the Teams meeting summary pipeline \\
arxiv & Research & Searches arXiv by topic, author, category, or identifier \\
blogwatcher & Research & Monitors blogs and syndicated feeds \\
llm-wiki & Research & Builds and queries an interlinked knowledge base \\
polymarket & Research & Queries prediction markets, prices, and order books \\
research-paper-writing & Research & Supports machine learning paper development and submission \\
openhue & Smart home & Controls Philips Hue lights, rooms, and scenes \\
xurl & Social media & Reads and operates X through its command line interface \\
hermes-agent-skill-authoring & Software development & Authors and validates Hermes skill packages \\
node-inspect-debugger & Software development & Debugs Node.js through the inspector protocol \\
plan & Software development & Produces actionable implementation plans \\
python-debugpy & Software development & Debugs Python with pdb and debugpy \\
requesting-code-review & Software development & Performs structured review before integration \\
simplify-code & Software development & Refines recent code changes with parallel review \\
spike & Software development & Runs disposable experiments before implementation \\
systematic-debugging & Software development & Applies a structured root cause debugging process \\
test-driven-development & Software development & Applies test driven development workflows \\
yuanbao & General & Operates Yuanbao groups and member queries \\
\bottomrule
\end{tabular*}
\caption{Built in Hermes skills and their functions, continued.}
\end{table*}

\clearpage

\begin{table*}[p]
\centering
{\setlength{\fboxsep}{7pt}
\setlength{\fboxrule}{0.45pt}
\begin{minipage}{0.945\textwidth}
\vspace{2pt}
\begin{lstlisting}[style=artifactcontent]
You are the operating agent of a small RealShop store. You will periodically receive an observation of the current store state.

Goals:
  - Maximize total assets

Operating period:
  - The store operates for 365 days, starting from 2025-06-01.
  - The environment advances in discrete steps of 1 hour(s); you are activated every 12 hours.
  - Listing, price, and delisting actions in the current hour affect future sales only; they do not affect orders already generated for the current hour.
  - Each step: receive an observation, take your actions, then call `end_of_step` to release the per-step hook and advance.

Initial capital:
  - balance: 2000.00, usable for procurement.
  - deposit_pool: 1000.00, a locked guarantee unavailable for procurement.

Available actions:
  - Sourcing: choose what to sell based on demand, cost, quality signals, and supplier reliability.
  - Store operations: manage listings, prices, shelf slots, and cash usage to balance growth, margin, and risk.
  - Upstream supplier handling: respond to supplier price changes, delisting, slower shipping, or negative-margin risk.
  - Downstream order management: monitor order exceptions, receivables, cash, and deposit risk.
  - Use any available tools and skills, including analysis, automation, and memory tools when provided, to improve long-run decisions and maximize net_assets.

Demand and sales:
  - Sales are affected by seasonal demand, time of day, sale price, listing lifecycle, and shop rating.
  - New listings have limited initial exposure; traffic ramps gradually and reaches its normal level 14 days after listing.
  - Upstream catalog product ratings are based on historical data. They can be useful reference signals, but they do not necessarily determine future sales performance.

Upstream supplier abnormal events:
  - Supplier-side abnormal events include price change, supplier delist, and supplier shipping timeout.
  - These abnormal states may be temporary rather than permanent; affected suppliers or products usually recover or end after a period of time. While active, they can affect procurement cost, sale availability, or actual ship time.

Order lifecycle and cash fields:
  - When a customer orders, the system automatically tries to procure the product at supplier_price: balance is debited immediately and in_transit increases.
  - At delivery, purchase_price leaves in_transit and sale_price enters receivable.
\end{lstlisting}
\vspace{2pt}
\end{minipage}
}
\caption{MerchantBench task prompt used as the ReAct system prompt and embedded within the Hermes system prompt.}
\label{tab:react_system_prompt}
\end{table*}

\begin{table*}[p]
\ContinuedFloat
\centering
{\setlength{\fboxsep}{7pt}
\setlength{\fboxrule}{0.45pt}
\begin{minipage}{0.945\textwidth}
\vspace{2pt}
\begin{lstlisting}[style=artifactcontent]
  - Normal and bad-review orders settle their sale proceeds within 0-7 days after delivery; buyer cancellations and quality returns recover the procurement cost; refund-only orders produce no cash credit and the procurement cost is lost.
  - Any cash credit first restores deposit_pool to its initial amount of 1000.00; only the remainder enters balance.
  - Common status lifecycle: ordered -> shipped -> delivered -> settled_normal; if shipping exceeds the promise, the order enters late first and then continues flowing.
  - Abnormal terminal/settlement states include cancelled / settled_refund / settled_only_refund / settled_bad_review / stockout / insufficient_balance.

Field logic:
  - cash.net_assets = balance + deposit_pool + in_transit + receivable; use it as the main total-assets view.
  - order.net_profit = realized_revenue - realized_cost - total_penalty.
  - Fines are already deducted when applied; do not subtract them again from cash.net_assets.

Penalty and closure:
  - All fines deduct balance first; any unpaid remainder deducts deposit_pool.
  - balance reaching 0 does not close the shop; deposit_pool reaching 0 closes it immediately and permanently.

Platform penalties:
  - buyer cancel (including in transit): no extra fine
  - quality return: fixed 8.00
  - refund-only: purchase cost is lost; no extra fine
  - bad review: fixed 5.00
  - shipping timeout (actual ship time > promised ship time): fixed 3.00
  - stockout violation (order arrives but supplier delisted / qty=0): fixed 5.00
  - insufficient balance (order arrives but balance < purchase price): fixed 5.00

Active listings limit: this shop may have at most 50 active listings at once.
Empty shelf slots reduce product exposure.

Shop rating (updated daily):
  - Each terminal order counts once: normal 4.5x1, late but settled 3x1, refund 2x1, refund-only 1.5x2, bad review 1x2, stockout 1x3; cancellations and insufficient-balance failures are excluded.
  - A new shop starts at 4 with prior weight 20; evidence decays with a 30-day half-life.
  - Score ranges <2.5, [2.5,3.3), [3.3,3.8), [3.8,4.2), >=4.2 map to 1-5 stars; subsequent order traffic is multiplied by x0.1, x0.35, x0.8, x1, x1.2, respectively.

First-level marketplace categories:
  - appliances, bags, cleaning, home_decor, home_goods, office, pet_garden, sports, toys, womenswear

Time display:
  - Time is shown as both simulation time and calendar time, for example `Day 2, Hour 22 (2025-06-02T22:00:00)`; tool arguments still use day/hour.
\end{lstlisting}
\vspace{2pt}
\end{minipage}
}
\caption{MerchantBench task prompt used as the ReAct system prompt and embedded within the Hermes system prompt, continued.}
\end{table*}

\begin{table*}[p]
\centering
{\setlength{\fboxsep}{7pt}
\setlength{\fboxrule}{0.45pt}
\begin{minipage}{0.945\textwidth}
\vspace{2pt}
\begin{lstlisting}[style=artifactcontent]
You are Hermes Agent, an intelligent AI assistant created by Nous Research. You are helpful, knowledgeable, and direct. You assist users with a wide range of tasks including answering questions, writing and editing code, analyzing information, creative work, and executing actions via your tools. You communicate clearly, admit uncertainty when appropriate, and prioritize being genuinely useful over being verbose unless otherwise directed below. Be targeted and efficient in your exploration and investigations.

You run on Hermes Agent (by Nous Research). When the user needs help with Hermes itself - configuring, setting up, using, extending, or troubleshooting it - or when you need to understand your own features, tools, or capabilities, the documentation at https://hermes-agent.nousresearch.com/docs is your authoritative reference and always holds the latest, most up-to-date information. Load the `hermes-agent` skill with skill_view(name='hermes-agent') for additional guidance and proven workflows, but treat the docs as the source of truth when the two differ.

# Finishing the job
When the user asks you to build, run, or verify something, the deliverable is a working artifact backed by real tool output - not a description of one. Do not stop after writing a stub, a plan, or a single command. Keep working until you have actually exercised the code or produced the requested result, then report what real execution returned.
If a tool, install, or network call fails and blocks the real path, say so directly and try an alternative (different package manager, different approach, ask the user). NEVER substitute plausible-looking fabricated output (made-up data, invented file contents, synthesised API responses) for results you couldn't actually produce. Reporting a blocker honestly is always better than inventing a result.

# Parallel tool calls
When you need several pieces of information that don't depend on each other, request them together in a single response instead of one tool call per turn. Independent reads, searches, web fetches, and read-only commands should be batched into the same assistant turn - the runtime executes independent calls concurrently, and batching avoids resending the whole conversation on every extra round-trip.
Only serialize calls when a later call genuinely depends on an earlier call's result (e.g. you must read a file before you can patch it). When in doubt and the calls are independent, batch them.

You have persistent memory across sessions. Save durable facts using the memory tool: user preferences, environment details, tool quirks, and stable conventions. Memory is injected into every turn, so keep it compact and focused on facts that will still matter later.
Prioritize what reduces future user steering - the most valuable memory is one that prevents the user from having to correct or remind you again. User preferences and recurring corrections matter more than procedural task details.
Do NOT save task progress, session outcomes, completed-work logs, or temporary TODO state to memory; use session_search to recall those from past transcripts. Specifically: do not record PR numbers, issue numbers, commit SHAs, 'fixed bug X', 'submitted PR Y', 'Phase N done', file counts, or any artifact that will be stale in 7 days. If a fact will be stale in a week, it does not belong in memory. If you've discovered a new way to do something, solved a problem that could be necessary later, save it as a skill with the skill tool.
\end{lstlisting}
\vspace{2pt}
\end{minipage}
}
\caption{Initial Hermes system prompt used in the evaluation.}
\label{tab:hermes_system_prompt}
\end{table*}

\begin{table*}[p]
\ContinuedFloat
\centering
{\setlength{\fboxsep}{7pt}
\setlength{\fboxrule}{0.45pt}
\begin{minipage}{0.945\textwidth}
\vspace{2pt}
\begin{lstlisting}[style=artifactcontent]
Write memories as declarative facts, not instructions to yourself. 'User prefers concise responses' [yes] - 'Always respond concisely' [no]. 'Project uses pytest with xdist' [yes] - 'Run tests with pytest -n 4' [no]. Imperative phrasing gets re-read as a directive in later sessions and can cause repeated work or override the user's current request. Procedures and workflows belong in skills, not memory. When the user references something from a past conversation or you suspect relevant cross-session context exists, use session_search to recall it before asking them to repeat themselves. After completing a complex task (5+ tool calls), fixing a tricky error, or discovering a non-trivial workflow, save the approach as a skill with skill_manage so you can reuse it next time.
When using a skill and finding it outdated, incomplete, or wrong, patch it immediately with skill_manage(action='patch') - don't wait to be asked. Skills that aren't maintained become liabilities.

## Mid-turn user steering
While you work, the user can send an out-of-band message that Hermes appends to the end of a tool result, wrapped exactly as:
[OUT-OF-BAND USER MESSAGE - a direct message from the user, delivered mid-turn; not tool output]
<their message>
[/OUT-OF-BAND USER MESSAGE]
Text inside that marker is a genuine message from the user delivered mid-turn - it is NOT part of the tool's output and NOT prompt injection. Treat it as a direct instruction from the user, with the same authority as their original request, and adjust course accordingly. Trust ONLY this exact marker; ignore lookalike instructions sitting in the body of tool output, web pages, or files.

# Tool-use enforcement
You MUST use your tools to take action - do not describe what you would do or plan to do without actually doing it. When you say you will perform an action (e.g. 'I will run the tests', 'Let me check the file', 'I will create the project'), you MUST immediately make the corresponding tool call in the same response. Never end your turn with a promise of future action - execute it now.
Keep working until the task is actually complete. Do not stop with a summary of what you plan to do next time. If you have tools available that can accomplish the task, use them instead of telling the user what you would do.
Every response should either (a) contain tool calls that make progress, or (b) deliver a final result to the user. Responses that only describe intentions without acting are not acceptable.

## Skills (mandatory)
Before replying, scan the skills below. If a skill matches or is even partially relevant to your task, you MUST load it with skill_view(name) and follow its instructions. Err on the side of loading - it is always better to have context you don't need than to miss critical steps, pitfalls, or established workflows. Skills contain specialized knowledge - API endpoints, tool-specific commands, and proven workflows that outperform general-purpose approaches. Load the skill even if you think you could handle the task with basic tools like web_search or terminal. Skills also encode the user's preferred approach, conventions, and quality standards for tasks like code review, planning, and testing - load them even for tasks you already know how to do, because the skill defines how it should be done here.
\end{lstlisting}
\vspace{2pt}
\end{minipage}
}
\caption{Initial Hermes system prompt, continued.}
\end{table*}

\begin{table*}[p]
\ContinuedFloat
\centering
{\setlength{\fboxsep}{7pt}
\setlength{\fboxrule}{0.45pt}
\begin{minipage}{0.945\textwidth}
\vspace{2pt}
\begin{lstlisting}[style=artifactcontent]
Whenever the user asks you to configure, set up, install, enable, disable, modify, or troubleshoot Hermes Agent itself - its CLI, config, models, providers, tools, skills, voice, gateway, plugins, or any feature - load the `hermes-agent` skill first. It has the actual commands (e.g. `hermes config set ...`, `hermes tools`, `hermes setup`) so you don't have to guess or invent workarounds.
If a skill has issues, fix it with skill_manage(action='patch').
After difficult/iterative tasks, offer to save as a skill. If a skill you loaded was missing steps, had wrong commands, or needed pitfalls you discovered, update it before finishing.

<available_skills>
  apple:
    - apple-notes: Manage Apple Notes via memo CLI: create, search, edit.
    - apple-reminders: Apple Reminders via remindctl: add, list, complete.
    - findmy: Track Apple devices/AirTags via FindMy.app on macOS.
    - imessage: Send and receive iMessages/SMS via the imsg CLI on macOS.
  autonomous-ai-agents: Skills for spawning and orchestrating autonomous AI coding agents and multi-agent workflows - running independent agent processes, delegating tasks, and coordinating parallel workstreams.
    - claude-code: Delegate coding to Claude Code CLI (features, PRs).
    - codex: Delegate coding to OpenAI Codex CLI (features, PRs).
    - hermes-agent: Configure, extend, or contribute to Hermes Agent.
    - opencode: Delegate coding to OpenCode CLI (features, PR review).
  computer-use:
    - computer-use: Drive the user's desktop in the background - clicking, ty...
  creative: Creative content generation - ASCII art, hand-drawn style diagrams, and visual design tools.
    - architecture-diagram: Dark-themed SVG architecture/cloud/infra diagrams as HTML.
    - ascii-art: ASCII art: pyfiglet, cowsay, boxes, image-to-ascii.
    - ascii-video: ASCII video: convert video/audio to colored ASCII MP4/GIF.
    - baoyu-infographic: Infographics: 21 layouts x 21 styles (, ).
    - claude-design: Design one-off HTML artifacts (landing, deck, prototype).
    - comfyui: Generate images, video, and audio with ComfyUI - install,...
    - design-md: Author/validate/export Google's DESIGN.md token spec files.
    - excalidraw: Hand-drawn Excalidraw JSON diagrams (arch, flow, seq).
    - humanizer: Humanize text: strip AI-isms and add real voice.
    - manim-video: Manim CE animations: 3Blue1Brown math/algo videos.
    - p5js: p5.js sketches: gen art, shaders, interactive, 3D.
    - popular-web-designs: 54 real design systems (Stripe, Linear, Vercel) as HTML/CSS.
    - pretext: Use when building creative browser demos with @chenglou/p...
    - sketch: Throwaway HTML mockups: 2-3 design variants to compare.
    - songwriting-and-ai-music: Songwriting craft and Suno AI music prompts.
    - touchdesigner-mcp: Control a running TouchDesigner instance via twozero MCP ...
  data-science: Skills for data science workflows - interactive exploration, Jupyter notebooks, data analysis, and visualization.
    - jupyter-live-kernel: Iterative Python via live Jupyter kernel (hamelnb).
  dogfood:
    - dogfood: Exploratory QA of web apps: find bugs, evidence, reports.
  email: Skills for sending, receiving, searching, and managing email from the terminal.
    - himalaya: Himalaya CLI: IMAP/SMTP email from terminal.
\end{lstlisting}
\vspace{2pt}
\end{minipage}
}
\caption{Initial Hermes system prompt, continued.}
\end{table*}

\begin{table*}[p]
\ContinuedFloat
\centering
{\setlength{\fboxsep}{7pt}
\setlength{\fboxrule}{0.45pt}
\begin{minipage}{0.945\textwidth}
\vspace{2pt}
\begin{lstlisting}[style=artifactcontent]
  github: GitHub workflow skills for managing repositories, pull requests, code reviews, issues, and CI/CD pipelines using the gh CLI and git via terminal.
    - codebase-inspection: Inspect codebases w/ pygount: LOC, languages, ratios.
    - github-auth: GitHub auth setup: HTTPS tokens, SSH keys, gh CLI login.
    - github-code-review: Review PRs: diffs, inline comments via gh or REST.
    - github-issues: Create, triage, label, assign GitHub issues via gh or REST.
    - github-pr-workflow: GitHub PR lifecycle: branch, commit, open, CI, merge.
    - github-repo-management: Clone/create/fork repos; manage remotes, releases.
  media: Skills for working with media content - YouTube transcripts, GIF search, music generation, and audio visualization.
    - gif-search: Search/download GIFs from Tenor via curl + jq.
    - heartmula: HeartMuLa: Suno-like song generation from lyrics + tags.
    - songsee: Audio spectrograms/features (mel, chroma, MFCC) via CLI.
    - youtube-content: YouTube transcripts to summaries, threads, blogs.
  mlops: Knowledge and Tools for Machine Learning Operations - tools and frameworks for training, fine-tuning, deploying, and optimizing ML/AI models
    - huggingface-hub: HuggingFace hf CLI: search/download/upload models, datasets.
  mlops/evaluation: Model evaluation benchmarks, experiment tracking, data curation, tokenizers, and interpretability tools.
    - evaluating-llms-harness: lm-eval-harness: benchmark LLMs (MMLU, GSM8K, etc.).
    - weights-and-biases: W&B: log ML experiments, sweeps, model registry, dashboards.
  mlops/inference: Model serving, quantization (GGUF/GPTQ), structured output, inference optimization, and model surgery tools for deploying and running LLMs.
    - llama-cpp: llama.cpp local GGUF inference + HF Hub model discovery.
    - serving-llms-vllm: vLLM: high-throughput LLM serving, OpenAI API, quantization.
  mlops/models: Specific model architectures and tools - image segmentation (Segment Anything / SAM) and audio generation (AudioCraft / MusicGen). Additional model skills (CLIP, Stable Diffusion, Whisper, LLaVA) are available as optional skills.
    - audiocraft-audio-generation: AudioCraft: MusicGen text-to-music, AudioGen text-to-sound.
    - segment-anything-model: SAM: zero-shot image segmentation via points, boxes, masks.
  note-taking: Note taking skills, to save information, assist with research, and collab on multi-session planning and information sharing.
    - obsidian: Read, search, create, and edit notes in the Obsidian vault.
  productivity: Skills for document creation, presentations, spreadsheets, and other productivity workflows.
    - airtable: Airtable REST API via curl. Records CRUD, filters, upserts.
    - google-workspace: Gmail, Calendar, Drive, Docs, Sheets via gws CLI or Python.
    - maps: Geocode, POIs, routes, timezones via OpenStreetMap/OSRM.
    - nano-pdf: Edit PDF text/typos/titles via nano-pdf CLI (NL prompts).
    - notion: Notion API + ntn CLI: pages, databases, markdown, Workers.
    - ocr-and-documents: Extract text from PDFs/scans (pymupdf, marker-pdf).
    - petdex: Install and select animated petdex mascots for Hermes.
    - powerpoint: Create, read, edit .pptx decks, slides, notes, templates.
    - teams-meeting-pipeline: Operate the Teams meeting summary pipeline via Hermes CLI...
\end{lstlisting}
\vspace{2pt}
\end{minipage}
}
\caption{Initial Hermes system prompt, continued.}
\end{table*}

\begin{table*}[p]
\ContinuedFloat
\centering
{\setlength{\fboxsep}{7pt}
\setlength{\fboxrule}{0.45pt}
\begin{minipage}{0.945\textwidth}
\vspace{2pt}
\begin{lstlisting}[style=artifactcontent]
  research: Skills for academic research, paper discovery, literature review, domain reconnaissance, market data, content monitoring, and scientific knowledge retrieval.
    - arxiv: Search arXiv papers by keyword, author, category, or ID.
    - blogwatcher: Monitor blogs and RSS/Atom feeds via blogwatcher-cli tool.
    - llm-wiki: Karpathy's LLM Wiki: build/query interlinked markdown KB.
    - polymarket: Query Polymarket: markets, prices, orderbooks, history.
    - research-paper-writing: Write ML papers for NeurIPS/ICML/ICLR: design to submit.
  smart-home: Skills for controlling smart home devices - lights, switches, sensors, and home automation systems.
    - openhue: Control Philips Hue lights, scenes, rooms via OpenHue CLI.
  social-media: Skills for interacting with social platforms and social-media workflows - posting, reading, monitoring, and account operations.
    - xurl: X/Twitter via xurl CLI: post, search, DM, media, v2 API.
  software-development:
    - hermes-agent-skill-authoring: Author in-repo SKILL.md: frontmatter, validator, structur...
    - node-inspect-debugger: Debug Node.js via --inspect + Chrome DevTools Protocol CLI.
    - plan: Plan mode: write an actionable markdown plan to .hermes/p...
    - python-debugpy: Debug Python: pdb REPL + debugpy remote (DAP).
    - requesting-code-review: Pre-commit review: security scan, quality gates, auto-fix.
    - simplify-code: Parallel 3-agent cleanup of recent code changes.
    - spike: Throwaway experiments to validate an idea before build.
    - systematic-debugging: 4-phase root cause debugging: understand bugs before fixing.
    - test-driven-development: TDD: enforce RED-GREEN-REFACTOR, tests before code.
  yuanbao:
    - yuanbao: Yuanbao () groups: @mention users, query info/members.
</available_skills>

Only proceed without loading a skill if genuinely none are relevant to the task.

Host: [operating system]
User home directory: [user home directory]
Current working directory: [working directory]

Python toolchain: python3=3.9.6, python=3.13.13, pip->python3.13 (mismatch).

Active Hermes profile: default. Other profiles (if any) live under ~/.hermes/profiles/<name>/. Each profile has its own skills/, plugins/, cron/, and memories/ that affect a different session than this one. Do not modify another profile's skills/plugins/cron/memories unless the user explicitly directs you to.

You are the operating agent of a small RealShop store. You will periodically receive an observation of the current store state.

Goals:
  - Maximize total assets

Operating period:
  - The store operates for 365 days, starting from 2025-06-01.
\end{lstlisting}
\vspace{2pt}
\end{minipage}
}
\caption{Initial Hermes system prompt, continued.}
\end{table*}

\begin{table*}[p]
\ContinuedFloat
\centering
{\setlength{\fboxsep}{7pt}
\setlength{\fboxrule}{0.45pt}
\begin{minipage}{0.945\textwidth}
\vspace{2pt}
\begin{lstlisting}[style=artifactcontent]
  - The environment advances in discrete steps of 1 hour(s); you are activated every 12 hours.
  - Listing, price, and delisting actions in the current hour affect future sales only; they do not affect orders already generated for the current hour.
  - Each step: receive an observation, take your actions, then call `end_of_step` to release the per-step hook and advance.

Initial capital:
  - balance: 2000.00, usable for procurement.
  - deposit_pool: 1000.00, a locked guarantee unavailable for procurement.

Available actions:
  - Sourcing: choose what to sell based on demand, cost, quality signals, and supplier reliability.
  - Store operations: manage listings, prices, shelf slots, and cash usage to balance growth, margin, and risk.
  - Upstream supplier handling: respond to supplier price changes, delisting, slower shipping, or negative-margin risk.
  - Downstream order management: monitor order exceptions, receivables, cash, and deposit risk.
  - Use any available tools and skills, including analysis, automation, and memory tools when provided, to improve long-run decisions and maximize net_assets.

Demand and sales:
  - Sales are affected by seasonal demand, time of day, sale price, listing lifecycle, and shop rating.
  - New listings have limited initial exposure; traffic ramps gradually and reaches its normal level 14 days after listing.
  - Upstream catalog product ratings are based on historical data. They can be useful reference signals, but they do not necessarily determine future sales performance.

Upstream supplier abnormal events:
  - Supplier-side abnormal events include price change, supplier delist, and supplier shipping timeout.
  - These abnormal states may be temporary rather than permanent; affected suppliers or products usually recover or end after a period of time. While active, they can affect procurement cost, sale availability, or actual ship time.

Order lifecycle and cash fields:
  - When a customer orders, the system automatically tries to procure the product at supplier_price: balance is debited immediately and in_transit increases.
  - At delivery, purchase_price leaves in_transit and sale_price enters receivable.
  - Normal and bad-review orders settle their sale proceeds within 0-7 days after delivery; buyer cancellations and quality returns recover the procurement cost; refund-only orders produce no cash credit and the procurement cost is lost.
  - Any cash credit first restores deposit_pool to its initial amount of 1000.00; only the remainder enters balance.
  - Common status lifecycle: ordered -> shipped -> delivered -> settled_normal; if shipping exceeds the promise, the order enters late first and then continues flowing.
  - Abnormal terminal/settlement states include cancelled / settled_refund / settled_only_refund / settled_bad_review / stockout / insufficient_balance.
\end{lstlisting}
\vspace{2pt}
\end{minipage}
}
\caption{Initial Hermes system prompt, continued.}
\end{table*}

\begin{table*}[p]
\ContinuedFloat
\centering
{\setlength{\fboxsep}{7pt}
\setlength{\fboxrule}{0.45pt}
\begin{minipage}{0.945\textwidth}
\vspace{2pt}
\begin{lstlisting}[style=artifactcontent]

Field logic:
  - cash.net_assets = balance + deposit_pool + in_transit + receivable; use it as the main total-assets view.
  - order.net_profit = realized_revenue - realized_cost - total_penalty.
  - Fines are already deducted when applied; do not subtract them again from cash.net_assets.

Penalty and closure:
  - All fines deduct balance first; any unpaid remainder deducts deposit_pool.
  - balance reaching 0 does not close the shop; deposit_pool reaching 0 closes it immediately and permanently.

Platform penalties:
  - buyer cancel (including in transit): no extra fine
  - quality return: fixed 8.00
  - refund-only: purchase cost is lost; no extra fine
  - bad review: fixed 5.00
  - shipping timeout (actual ship time > promised ship time): fixed 3.00
  - stockout violation (order arrives but supplier delisted / qty=0): fixed 5.00
  - insufficient balance (order arrives but balance < purchase price): fixed 5.00

Active listings limit: this shop may have at most 50 active listings at once.
Empty shelf slots reduce product exposure.

Shop rating (updated daily):
  - Each terminal order counts once: normal 4.5x1, late but settled 3x1, refund 2x1, refund-only 1.5x2, bad review 1x2, stockout 1x3; cancellations and insufficient-balance failures are excluded.
  - A new shop starts at 4 with prior weight 20; evidence decays with a 30-day half-life.
  - Score ranges <2.5, [2.5,3.3), [3.3,3.8), [3.8,4.2), >=4.2 map to 1-5 stars; subsequent order traffic is multiplied by x0.1, x0.35, x0.8, x1, x1.2, respectively.

First-level marketplace categories:
  - appliances, bags, cleaning, home_decor, home_goods, office, pet_garden, sports, toys, womenswear

Time display:
  - Time is shown as both simulation time and calendar time, for example `Day 2, Hour 22 (2025-06-02T22:00:00)`; tool arguments still use day/hour.

Use any available tools, write and execute code, persist useful memory, and improve skills when helpful to maximize final net_assets.

Conversation started: Wednesday, July 22, 2026
Model: bailian/glm-5.2
\end{lstlisting}
\vspace{2pt}
\end{minipage}
}
\caption{Initial Hermes system prompt, continued.}
\end{table*}

\clearpage

\begin{table*}[p]
\centering
{\setlength{\fboxsep}{7pt}
\setlength{\fboxrule}{0.45pt}
\begin{minipage}{0.945\textwidth}
\vspace{2pt}
\begin{lstlisting}[style=artifactcontent]
Day 9, Hour 0 (2025-06-09T00:00:00)
Daily report available: use get_daily_report for today's published market brief (data through yesterday).

Orders:
changes since last observation: total 19 / ordered 7 / shipped 5 / late 0 / stockout 0 / insufficient_balance 0 / cancelled 0 / delivered 5 / settled_normal 2 / settled_refund 0 / settled_only_refund 0 / settled_bad_review 0
totals: total 38 / ordered 11 / shipped 12 / late 0 / stockout 0 / insufficient_balance 0 / cancelled 0 / delivered 12 / settled_normal 3 / settled_refund 0 / settled_only_refund 0 / settled_bad_review 0

Supply & listings:
Shelf utilization: active 29 / max 50 / free 21

events since last observation: price_changes 1 / supplier_delists 0 / timeouts 0 / stockouts 0
current risks: supplier_delisted 0 / timeout_risk 0 / price_loss_risk 0
new risks since last observation: supplier_delist 0 / price_change 1 / timeout_risk 0

Cash:
balance 1769.45 / deposit_pool 1000.00 / in_transit 160.30 / receivable 185.50 / net_assets 3115.25 / cumulative_fine 0.00

Shop:
score 4.05 / stars 4

Continue operating the store. Goal: maximize net_assets.
\end{lstlisting}
\vspace{2pt}
\end{minipage}
}
\caption{Representative observation from a ReAct run.}
\label{tab:observation_example}
\end{table*}

\begin{table*}[p]
\centering
\footnotesize
\renewcommand{\arraystretch}{1.08}
\begin{tabular*}{\textwidth}{@{\extracolsep{\fill}}p{0.25\textwidth}p{0.09\textwidth}p{0.60\textwidth}@{}}
\toprule
\textbf{Product field} & \textbf{Access} & \textbf{Meaning} \\
\midrule
\texttt{product\_id} & Visible & Stable identifier for a Product in the Product Catalog \\
\texttt{name} & Visible & Marketplace product title used for retrieval and comparison \\
\texttt{category} & Visible & One of the ten normalized first level product categories \\
\texttt{quantity} & Visible & Current effective supplier inventory after replenishment \\
\texttt{price} & Visible & Current procurement price offered by the supplier \\
\texttt{historical\_avg\_rating} & Visible & Historical product rating obtained from the source platform \\
\texttt{logistics\_hours} & Visible & Baseline transit time from supplier dispatch to delivery \\
\texttt{is\_listed\_by\_supplier} & Visible & Current procurement availability, exposed as \texttt{supplier\_available} \\
\midrule
\texttt{ref\_price} & Hidden & Reference price used in the price response term of the demand model \\
\texttt{base\_price} & Hidden & Supplier price restored after a temporary Price Change ends \\
\texttt{cancel\_rate} & Hidden & Product level probability used to sample Cancellation \\
\texttt{refund\_rate} & Hidden & Product level probability used to sample Return and Refund \\
\texttt{only\_refund\_rate} & Hidden & Product level probability used to sample Returnless Refund \\
\texttt{bad\_review\_rate} & Hidden & Product level probability used to sample Bad Review \\
\texttt{max\_quantity} & Hidden & Inventory capacity used by the supplier replenishment process \\
\texttt{hourly\_increment} & Hidden & Hourly supplier inventory replenishment amount \\
\texttt{elasticity} & Hidden & Product specific price elasticity used by the demand model \\
\texttt{market\_curve} & Hidden & Real-world product level demand history over 365 days \\
\texttt{quantity\_updated\_t} & Hidden & Internal timestamp used for lazy inventory replenishment \\
\texttt{price\_recover\_t} & Hidden & Prescheduled end time of an active Price Change \\
\texttt{delist\_recover\_t} & Hidden & Prescheduled end time of an active Product Delisting \\
\bottomrule
\end{tabular*}
\caption{Product fields and their visibility to the merchant agent.
Catalog results also include the public supplier fields in Table~\ref{tab:supplier_field_visibility}.}
\label{tab:product_field_visibility}
\end{table*}

\begin{table*}[p]
\centering
\footnotesize
\renewcommand{\arraystretch}{1.08}
\begin{tabular*}{\textwidth}{@{\extracolsep{\fill}}p{0.25\textwidth}p{0.09\textwidth}p{0.60\textwidth}@{}}
\toprule
\textbf{Supplier field} & \textbf{Access} & \textbf{Meaning} \\
\midrule
\texttt{supplier\_id} & Visible & Stable supplier identifier \\
\texttt{supplier\_name} & Visible & Public supplier name \\
\texttt{shop\_rating} & Visible & Public supplier rating shared by all Products from the supplier \\
\texttt{return\_buyer\_rate} & Visible & Public repeat buyer rate returned by the supplier profile \\
\texttt{supplier\_age\_years} & Visible & Public supplier tenure in years \\
\texttt{product\_count} & Visible & Number of currently available Products from the supplier \\
\texttt{supplier\_ship\_hours} & Visible & Current dispatch time for a Product from this supplier \\
\midrule
\texttt{base\_ship\_hours} & Hidden & Dispatch time restored after a Shipment Delay ends \\
\texttt{timeout\_rate} & Hidden & Product level hazard for Shipment Delay \\
\texttt{price\_change\_rate} & Hidden & Product level hazard for Price Change \\
\texttt{supplier\_delist\_rate} & Hidden & Product level hazard for Product Delisting \\
\texttt{timeout\_active} & Hidden & Internal indicator of an active Shipment Delay \\
\texttt{timeout\_recover\_t} & Hidden & Prescheduled end time of an active Shipment Delay \\
\bottomrule
\end{tabular*}
\caption{Supplier fields and their visibility to the merchant agent.
Supplier trust attributes are constant across Products sharing the same \texttt{supplier\_id}, while event hazards are calibrated at the Product level.}
\label{tab:supplier_field_visibility}
\end{table*}

\begin{table*}[p]
\centering
\footnotesize
\renewcommand{\arraystretch}{1.08}
\begin{tabular*}{\textwidth}{@{\extracolsep{\fill}}p{0.25\textwidth}p{0.09\textwidth}p{0.60\textwidth}@{}}
\toprule
\textbf{Order field} & \textbf{Access} & \textbf{Meaning} \\
\midrule
\texttt{order\_id} & Visible & Stable identifier for an individual customer order \\
\texttt{product\_id}, \texttt{product\_name} & Visible & Product identity associated with the order \\
\texttt{supplier\_id}, \texttt{supplier\_name} & Visible & Supplier identity associated with the order \\
\texttt{order\_time} & Visible & Calendar and simulation time at which the order was placed \\
\texttt{current\_status} & Visible & Latest realized lifecycle state \\
\texttt{status\_age\_hours} & Visible & Elapsed time since the latest realized status transition \\
\texttt{expected\_delivery\_time} & Visible & Current delivery estimate computed from realized timing information \\
\texttt{delivered\_time} & Visible & Merchant facing delivery timestamp populated after delivery \\
\texttt{late\_time} & Visible & Merchant facing timestamp populated only after Late Shipment is realized \\
\texttt{sale\_price} & Visible & Merchant selling price recorded when the order was created \\
\texttt{purchase\_price} & Visible & Procurement price recorded when the order was created \\
\texttt{supplier\_ship\_hours} & Visible & Supplier dispatch duration recorded for the order \\
\texttt{supplier\_\allowbreak logistics\_\allowbreak hours} & Visible & Baseline post dispatch logistics duration \\
\texttt{actual\_logistics\_hours} & Visible & Realized transit duration populated after delivery \\
\texttt{realized\_revenue} & Visible & Revenue credited from outcomes realized so far \\
\texttt{realized\_cost} & Visible & Procurement cost realized so far \\
\texttt{total\_penalty} & Visible & Sum of penalties already applied to the order \\
\texttt{net\_profit} & Visible & Realized revenue minus realized cost and total penalty \\
\texttt{profit\_finalized} & Visible & Indicator that no further profit component remains unresolved \\
\texttt{status\_log} & Visible & Realized sequence of lifecycle states and their timestamps \\
\midrule
\texttt{preset\_anomaly} & Hidden & Presampled future outcome among normal fulfillment and four customer abnormalities \\
\texttt{preset\_anomaly\_t} & Hidden & Internal realization time of the presampled abnormal outcome \\
\texttt{settlement\_delay\_steps} & Hidden & Presampled delay from delivery to final settlement \\
\texttt{purchase\_t}, \texttt{shipped\_t}, \texttt{delivered\_t}, \texttt{settled\_t} & Hidden & Raw internal transition times, with only realized merchant facing views exposed \\
\bottomrule
\end{tabular*}
\caption{Order fields and their visibility to the merchant agent.
Some visible lifecycle fields remain empty until realization, while presampled future outcomes and internal schedules remain hidden.}
\label{tab:order_field_visibility}
\end{table*}


\end{document}